\documentclass[conference,onecolumn]{IEEEtran}

\IEEEoverridecommandlockouts

\usepackage{cite}
\usepackage[margin=1in]{geometry}
\usepackage{amsmath,amssymb,amsfonts}
\usepackage{booktabs}
\usepackage{array}
\usepackage{tabularx}
\usepackage{enumitem}
\usepackage{graphicx}
\usepackage{xcolor}
\usepackage{listings}
\usepackage{hyperref}
\usepackage{url}
\usepackage{microtype}
\usepackage{tikz}
\usepackage{placeins}
\usepackage{fancyhdr}
\usetikzlibrary{arrows.meta,positioning,fit,calc}

\hypersetup{
  hidelinks,
  pdftitle={Intent-Governed Tool Authorization for AI Agents},
  pdfauthor={Genliang Zhu; Chu Wang},
  pdfsubject={Request-bounded authorization for tool-using AI agents},
  pdfkeywords={AI agents, authorization, access control, least privilege, prompt injection}
}

\setlist[itemize]{leftmargin=1.2em}
\setlist[enumerate]{leftmargin=1.4em}
\newcommand{\sys}{OpenPort}
\newcommand{\tool}{\mathcal{T}}
\newcommand{\visible}{\mathsf{Visible}}
\newcommand{\allow}{\mathsf{Allow}}
\newcommand{\effects}{\mathsf{Effects}}
\newcommand{\policy}{\mathcal{P}}

\newcommand{\risk}{\mathsf{Risk}}

\newcommand{\consistent}{\mathsf{Consistent}}
\newcommand{\mcode}[1]{\text{\texttt{#1}}}
\newcommand{\statementheading}[1]{%
  \par\smallskip\noindent\textbf{#1.}\par\nobreak\noindent}
\newcolumntype{L}[1]{>{\raggedright\arraybackslash}p{#1}}
\newcolumntype{Y}{>{\raggedright\arraybackslash}X}

\definecolor{opBlue}{HTML}{3B82F6}
\definecolor{opBlueFill}{HTML}{EFF6FF}
\definecolor{opBlueText}{HTML}{1E40AF}
\definecolor{opGreen}{HTML}{22C55E}
\definecolor{opGreenFill}{HTML}{F0FDF4}
\definecolor{opGreenText}{HTML}{15803D}
\definecolor{opAmber}{HTML}{FCD34D}
\definecolor{opAmberFill}{HTML}{FEF3C7}
\definecolor{opAmberText}{HTML}{92400E}
\definecolor{opRed}{HTML}{EF4444}
\definecolor{opRedFill}{HTML}{FEF2F2}
\definecolor{opRedText}{HTML}{DC2626}
\definecolor{opPurple}{HTML}{8B5CF6}
\definecolor{opPurpleFill}{HTML}{F5F3FF}
\definecolor{opPurpleText}{HTML}{7C3AED}

\newenvironment{floatbox}{\begin{minipage}{0.92\linewidth}\centering}{\end{minipage}}

\lstdefinelanguage{json}{
  morestring=[b]",
  showstringspaces=false,
  morecomment=[l]{//},
  morecomment=[s]{/*}{*/},
  keywords={true,false,null},
  sensitive=false
}

\title{\textbf{Intent-Governed Tool Authorization for AI Agents}}
\author{%
\parbox[t]{0.48\linewidth}{\centering
  Genliang Zhu\\
  Accentrust\\
  Georgia Institute of Technology
}
\hfill
\parbox[t]{0.48\linewidth}{\centering
  Chu Wang\\
  Accentrust\\
  University of Illinois Urbana-Champaign
}
\thanks{Correspondence: \href{mailto:research@accentrust.com}{research@accentrust.com}.}%
}

\begin{document}
\maketitle
\pagestyle{fancy}
\fancyhf{}
\renewcommand{\headrulewidth}{0pt}
\renewcommand{\footrulewidth}{0pt}
\fancyfoot[R]{\thepage}
\thispagestyle{fancy}

\begin{abstract}
Tool-using AI agents often use credentials broader than a user's current request. We present an \emph{intent-governed tool-authorization layer} that derives a short-lived certificate from a trusted request, narrows the authorized tool manifest, and checks proposed tool and payload effects before execution. It cannot exceed static policy; request confinement also requires certificate fidelity and sound effect bounds. We evaluate it over OpenPort with endpoint tests, 176 runtime-backed synthetic tasks, real-model pilots, 306 end-to-end trials, and a 36-trial benchmark-shaped external subset. In the deterministic comparison, reference certificates reduce the archived composite exposure-or-path indicator from 1.0000 to 0. Across end-to-end model runs, the combined path records no completed unsafe executions; unsafe accepted authority remains 0.0909--0.2727, entirely as non-executed drafts. Trace-backed normalization removes this residual at substantial utility cost. The results support static-policy non-expansion and identify certificate precision as the principal remaining bottleneck.
\end{abstract}


\section{Introduction}

AI agents are moving from conversational assistance to operational action. A deployed agent can inspect records, update state, export data, coordinate workflows, and act across application boundaries. In such settings, tool access is no longer a convenience feature of a language model interface. It is an application security boundary.

The access-control question normally asked by an application gateway is credential-centric: does this token, app, or integration have permission to call this endpoint? Credential-centric authorization is necessary. It supports revocation, least privilege, tenant isolation, and accountability. But agentic tool use introduces a second question that traditional API authorization rarely models: \emph{is the proposed tool call justified by what the user currently asked the agent to do?}

This distinction matters because the same credential can be legitimately broad while a particular user request is narrow. An agent may have both read and export scopes because the organization approves those capabilities for some workflows. Yet if the user asks ``summarize recent records for this period,'' exposing a bulk export tool to the model is unnecessary and potentially dangerous. If the model later proposes an export because an external document, retrieved note, or malicious tool description influenced it, credential-only authorization cannot distinguish the overbroad effect from the user's original summary request. Likewise, a model may construct a plausible update payload that touches a different resource than the user named, or a delete call that exceeds the intended target set.

Recent security research demonstrates this failure mode. AgentDojo evaluates prompt-injection attacks and defenses in tool-using agents~\cite{debenedetti2024agentdojo}. InjecAgent studies indirect prompt injection in tool-integrated agents and categorizes attacks into direct harm and data exfiltration~\cite{zhan2024injecagent}. ToolHijacker and ToolTweak show that tool descriptions and names can bias or hijack tool selection~\cite{shi2025toolhijacker,sneh2025tooltweak}. MCP-oriented work identifies tool poisoning, shadowing, and protocol-specific semantic attacks~\cite{wang2025mcptox,jamshidi2025securingmcp,huang2026mcptm,shen2026mcp38}. These works converge on a practical conclusion: the model's selected tool is insufficient evidence of authorization.

Industry guidance reaches a similar conclusion. OWASP lists prompt injection as the first 2025 LLM application risk and recommends least privilege, human approval for high-risk actions, external-content segregation, and adversarial testing~\cite{owasp2025llmtop10,owasp2025promptinjection}. NIST's Generative AI Profile recommends operational monitoring, assessment of embedded tools and APIs, and deployment-like assurance criteria~\cite{nist2024genaiprofile}. The 2025-06-18 Model Context Protocol (MCP) specification provides transport and tool-discovery semantics, including authorization guidance and model-controlled tools~\cite{mcp2025authorization,mcp2025tools}. Per-request authorization of a tool's intended effect remains outside these mechanisms.

This paper proposes an \emph{intent-certificate authorization layer} in front of the existing effect-control layer. It receives the user's request, derives an intent certificate, computes a session policy that can only narrow existing static authorization, filters the tool manifest accordingly, and checks that each proposed tool call and payload remains consistent with the certificate. Certificate-inconsistent proposals deny or clarify; certificate-consistent but high-risk proposals may enter draft, preflight, or confirmation routing. Unknown, low-confidence, or ambiguous requests cannot authorize immediate execution. The model, classifier, and planner are not trusted principals. They are advisors whose outputs are evaluated by the server-side policy layer.

\subsection{Relationship to OpenPort Protocol}

The OpenPort Protocol paper establishes a governance substrate for agent tool access: authorization-dependent discovery, scoped permissions, ABAC-like data policy constraints, draft-first writes, preflight impact hashing, optional state-witness checks, idempotency, rate limits, and audit~\cite{zhu2026openport}. We treat that prior work as the \emph{effect-control layer}. It determines whether a tool effect is permitted under static integration policy and risk-gated write semantics.

The intent-certificate layer supplies upstream request justification and composes with OpenPort Protocol:

\[
\begin{aligned}
\text{user intent}
&\rightarrow \text{dynamic least privilege}\\
&\rightarrow \text{filtered tools}
\rightarrow \text{governed effects}.
\end{aligned}
\]

The two layers address distinct decisions. OpenPort evaluates whether an effect is permitted by static integration policy and risk controls; the intent layer evaluates whether the current request justifies that effect. Together they combine request-level intent certificates with authoritative gateway enforcement.

\begin{table*}[t]
\centering
\footnotesize
\begin{tabularx}{\textwidth}{L{0.13\textwidth}L{0.16\textwidth}YYYY}
\toprule
\textbf{Mechanism} & \textbf{Authorization Unit} & \textbf{Per-Request Narrowing} & \textbf{Manifest Filtering} & \textbf{Payload-Effect Check} & \textbf{Planner Assumption} \\
\midrule
RBAC & Role & No & No & Limited & Orthogonal \\
ABAC & Attributes & Policy-defined & Policy-defined & Policy-defined & Orthogonal \\
OAuth scopes & Token/scope & No & No & No & Orthogonal \\
Usage control & Subject/object use & Policy-defined & Not intrinsic & Ongoing policy checks & Orthogonal \\
Caveat credentials & Credential + caveats & By attenuation & Not intrinsic & Caveat-defined & Orthogonal \\
Purpose-based AC & Declared purpose & Policy-defined & No & Partial & Orthogonal \\
MCP authorization & Resource server / token & Implementation-defined & Server-defined & No standard gate & Implementation-specific \\
Intent layer & Expressed task certificate & Yes & Yes & Yes & Untrusted \\
\bottomrule
\end{tabularx}
\caption{The intent layer adds request-bounded least privilege for untrusted agent planners on top of existing access-control families.}
\label{tab:noveltycompare}
\end{table*}

\subsection{Contributions}

This paper makes five contributions.

\begin{enumerate}
  \item \textbf{Problem formulation.} We define intent-tool mismatch as an access-control failure mode for AI agents: a proposed tool call may be statically authorized yet not authorized by the user's current expressed intent.
  \item \textbf{Formal model with explicit trust conditions.} We define intent certificates, monotone session policy, intent-aware manifests, consistency predicates, and least-privilege effect minimization. We separate unconditional static-policy non-expansion from request-level confinement, which additionally requires a sound certificate and effect-bound interpretation.
  \item \textbf{System design.} We design an intent-certificate gateway that narrows OpenPort manifests and action execution without trusting the model or classifier as final authority.
  \item \textbf{Implementation mapping.} We map the intent layer onto an OpenPort-style governance substrate and evaluate the resulting reference prototype.
  \item \textbf{Runtime-path evaluation.} We provide deterministic policy checks, endpoint-level runtime tests, a 176-task runtime-backed synthetic microbenchmark, a real-LLM classifier/planner pilot, an end-to-end LLM-in-the-loop runtime benchmark over 34 synthetic tasks with 3 repeats per model, and a benchmark-shaped runtime pilot over 12 adapted tasks and 3 models.
\end{enumerate}

\subsection{Intent Semantics}

The authorization layer uses an operational definition of ``intent'' tied to enforceable request content rather than a user's unobservable mental state:

\begin{itemize}
  \item \textbf{Expressed intent}: the user request as stated in trusted conversation input.
  \item \textbf{Inferred intent}: the classifier or parser's interpretation of that request.
  \item \textbf{Authorized intent}: the portion of the inferred request that is still permitted after static OpenPort policy, scopes, workspace guards, and risk policy are applied.
  \item \textbf{Operational intent certificate}: the bounded, auditable server-side record used by the authorization layer at enforcement time.
\end{itemize}

The operational certificate therefore represents a bounded interpretation of the expressed request. It may narrow static authority but never widen it.

\section{Background and Motivation}

\subsection{Tool Use as an Access-Control Boundary}

Tool-using agents mediate between natural language and application effects. In a conventional API client, the caller is usually deterministic software written by the application developer or an integration partner. In an agentic workflow, the immediate caller may be a planner driven by a language model, influenced by a conversation, untrusted documents, tool outputs, memory, retrieved snippets, or tool metadata. The agent may select tools automatically and may construct payloads that the user never directly inspected.

The MCP tools specification explicitly frames tools as model-controlled: models may discover and invoke tools automatically based on context and user prompts~\cite{mcp2025tools}. That design is useful for interoperability, but it shifts security burden to implementations. If a server exposes a tool list to a model, the server still needs to decide which tools should be visible in the current context and which proposed calls should be accepted.

Traditional access control offers relevant foundations. Least privilege limits authority~\cite{saltzer1975protection}. RBAC associates permissions with roles~\cite{hu1992rbac}. ABAC uses attributes of subjects, objects, actions, and environments to decide access~\cite{nist2014abac}. OAuth-style systems bind tokens and audiences, and MCP authorization guidance emphasizes validating tokens for the intended resource server~\cite{mcp2025authorization}. Usage control further supports obligations, environmental conditions, mutable attributes, and continuing decisions over an ongoing use~\cite{park2004ucon}; macaroon credentials support decentralized delegation whose authority can be attenuated with contextual caveats~\cite{birgisson2014macaroons}. These mechanisms can enforce a request bound once it is represented. The intent layer addresses the preceding agent-specific question: how to derive a bounded authorization envelope from the user's current trusted request and apply it both to the model-visible tool surface and to the proposed payload effect.

\subsection{Why Static Scopes Are Not Enough}

Static scopes describe what an integration may ever do, not what it should do for this request. Consider a tool set with:

\begin{itemize}
  \item \texttt{resource.read}, a read tool;
  \item \texttt{resource.export}, a high-risk export tool;
  \item \texttt{resource.create}, a write tool that creates a draft;
  \item \texttt{resource.delete}, a high-risk destructive tool.
\end{itemize}

An integration may legitimately receive all four scopes because an organization wants the agent to help with multiple workflows. But a specific request such as ``show records from the past period'' does not justify export or delete authority. A request such as ``create an entry for yesterday'' may justify a narrow create draft, but not arbitrary updates. A request such as ``delete the duplicate record'' justifies a high-risk delete candidate only under review and preflight constraints. Intent refines the static policy for the current session.

Scopes remain the upper bound, but they are insufficient as the sole authorization input for a probabilistic, context-sensitive caller. Intent-governed authorization makes the following rule explicit:

\[
\policy_{\text{session}}(u) \preceq \policy_{\text{static}}(a),
\]

where \(u\) is the user request, \(a\) is the integration app, and \(\preceq\) means ``no more permissive than.'' User intent narrows authority but never widens it.

\subsection{Prompt Injection and Tool Selection}

Prompt injection is a central motivation but not the only threat. OWASP describes direct and indirect prompt injection, including attacks through websites, files, images, and other external content, and notes impacts such as unauthorized function access and critical decision manipulation~\cite{owasp2025promptinjection}. AgentDojo and InjecAgent show that agents can be hijacked by untrusted tool outputs and external content~\cite{debenedetti2024agentdojo,zhan2024injecagent}. ToolHijacker, ToolTweak, MCPTox, and MCP threat-modeling work show that tool metadata itself can become an attack surface~\cite{shi2025toolhijacker,sneh2025tooltweak,wang2025mcptox,huang2026mcptm}.

The intent-certificate layer constrains effects even when malicious instructions influence the model. The server checks each proposed effect against certified intent and static policy, then hides the tool, denies the call, or routes it to draft/preflight review when the check fails. This shifts defense from model obedience to enforceable effect constraints.

\subsection{OpenPort as Substrate}

An OpenPort-style governance substrate provides the effect-control primitives that the intent layer requires. Such a substrate filters manifests by scopes, enforces policy constraints and workspace boundaries, creates drafts for writes, computes preflight impact hashes, optionally binds state witnesses, performs idempotency replay, and emits audit events.

The intent-certificate layer applies these effect-control primitives at the manifest and action boundaries for external agents.

\section{Running Examples}

\subsection{Bounded Read vs Bulk Export}

Suppose the integration has both \texttt{resource.read} and \texttt{resource.export} scopes. The user asks:

\begin{quote}
Show me this period's records and summarize any anomalies.
\end{quote}

A credential-only system may expose both tools. Under the intent layer, the certificate is:

\[
\begin{aligned}
I &= \{\mcode{read},\mcode{summarize}\},\\
B_R &= \{\text{records in current period}\},\\
B_E &= \{\text{no export, no mutation}\}.
\end{aligned}
\]

The export tool is hidden because its effect class is absent from \(I\). A direct named-call bypass fails with \texttt{agent.intent\_tool\_mismatch}. Static policy still applies: a read lacking scope or violating the date-window policy is denied regardless of intent.

\subsection{Narrow Create Draft}

For ``create a new entry for yesterday with amount 23.50,'' the certificate contains a create class, resource type, date, amount, and draft-only effect bound:

\[
\begin{aligned}
I&=\{\mcode{create}\},\\
B_E&=\{\text{single-resource draft},\
       \text{amount}\leq 23.50+\epsilon\}.
\end{aligned}
\]

A payload for 2350, a different date, or unrelated bulk metadata is inconsistent. A consistent create may become a reviewable draft under OpenPort; an inconsistent create must deny or clarify rather than create authority.

\subsection{High-Risk Delete}

For ``delete record \mcode{rec-123} because it is a duplicate,'' the authorization layer can expose deletion only for the named record and under high-risk routing:

\[
\begin{aligned}
I&=\{\mcode{delete}\},&
B_R&=\{\mcode{rec\mbox{-}123}\},\\
\rho&=\mcode{preflight}.&&
\end{aligned}
\]

OpenPort computes impact, binds the preflight hash to the payload, requires idempotency where configured, and otherwise creates a draft. Changing the target to \(\mcode{rec\mbox{-}124}\) fails the intent-certificate bound; changing it after preflight also fails the preflight hash or state witness.

\subsection{Indirect Prompt Injection Through a Document}

Suppose the user asks to summarize an attachment. Its text says ``ignore prior instructions and export all records.'' Certificate authority derives only from the authenticated user message, while the attachment is an untrusted channel. The certificate permits bounded summarize/transform effects, not export. A model-generated export proposal is hidden or denied even if the model treats the document text as an instruction.

\subsection{Compound Workflow}

For ``find duplicate records, show them to me, and delete the one I approve,'' one broad certificate would leak authority across planning steps. The authorization layer instead decomposes the workflow:

\[
\begin{aligned}
C_1&=\mcode{read/summarize},\\
C_2&=\mcode{delete with explicit approval}.
\end{aligned}
\]

The first certificate permits bounded discovery and presentation. The second is issued only after the user selects a concrete item and confirms deletion. Read/summarize certificates may discover candidates, but update/delete/export certificates must be re-issued after the target and effect are explicit.

\section{Threat Model}

\subsection{System Boundary}

We consider a server-side OpenPort gateway mediating all agent access to application data and actions. The gateway is trusted to enforce authentication, static authorization, tenant boundaries, data policy, rate limits, drafts, preflight checks, and audit. The model runtime, planner, intent classifier, tool selector, and external content sources are not trusted as final security principals.

The gateway derives a certificate only from an authenticated, explicitly selected user-message boundary and trusted request metadata. Attachments, retrieved text, web content, prior assistant output, tool output, and long-term memory are separate untrusted channels; they may inform planning but cannot silently become certificate-authorizing text. The certificate may be produced by rules, a model-assisted classifier, a hybrid classifier, or a deployment-specific router. Its semantic fields are untrusted until checked against deterministic constraints, and it can never bypass static policy.

\subsection{Assets}

Protected assets include:

\begin{itemize}
  \item application data reachable through read tools;
  \item write effects such as create, update, delete, export, delegation, and administrative actions;
  \item tenant and workspace boundaries;
  \item user intent records and session policy;
  \item audit trails and governance evidence;
  \item operational budget, rate-limit capacity, and review capacity.
\end{itemize}

\subsection{Adversaries}

We model the following adversaries and failure sources:

\begin{itemize}
  \item \textbf{Malicious user}: submits a prompt that tries to obtain broader tool access than policy permits.
  \item \textbf{External content attacker}: embeds instructions in webpages, files, emails, resumes, retrieved documents, or tool outputs.
  \item \textbf{Tool metadata attacker}: attempts to bias tool selection through malicious tool names, descriptions, or descriptors.
  \item \textbf{Compromised or overbroad integration}: holds credentials with more static scopes than needed for a specific request.
  \item \textbf{Benign model error}: selects the wrong tool, constructs an overbroad payload, retries unsafely, or drifts across turns.
  \item \textbf{Certificate attacker}: replays, swaps across actors/tenants, mutates, or supplies a certificate bound to stale policy or tool semantics.
  \item \textbf{Review-capacity attacker}: creates plausible drafts or confirmation requests to exhaust reviewers or induce approval fatigue.
  \item \textbf{Adapter attacker}: exploits a mismatch between declared and realized tool effects.
  \item \textbf{Operator misconfiguration}: grants broad scopes or weak auto-execute policy.
\end{itemize}

\subsection{Threats}

Table~\ref{tab:threats} summarizes the main threats.

\begin{table}[t]
\centering
\small
\begin{tabular}{L{0.21\linewidth}L{0.31\linewidth}L{0.32\linewidth}}
\toprule
\textbf{Threat} & \textbf{Failure mode} & \textbf{Authorization response} \\
\midrule
Intent-tool mismatch & Model proposes export/delete when user requested read/summary. & Hide high-risk tools; deny or clarify inconsistent calls. \\
Payload expansion & Tool class is plausible but payload touches resources or ranges not requested. & Check certificate bounds; deny or clarify on mismatch. \\
Intent drift & Multi-turn context shifts model plan away from original request. & Expire certificates; issue per-step leases; reclassify or require confirmation. \\
Indirect prompt injection & External content steers tool choice or certificate fields. & Separate trusted-message input; static non-expansion remains unconditional. \\
Tool metadata poisoning & Malicious descriptor biases selection. & Server filters selected tool through intent and static policy before execution. \\
Classifier error & Intent layer misclassifies request or emits weak bounds. & Certificate output cannot cross static policy, but may fail to narrow to the request; deterministic bounds and review contain the residual path. \\
Replay / binding swap & Certificate crosses actor, tenant, session, or policy version. & Bind subject and versions; expire/revoke; fail closed. \\
Review exhaustion & Attacker creates misleading draft volume. & Count drafts as authority; rate-limit and bind reviewer context. \\
Over-defense & Benign task denied unnecessarily. & Evaluate over-defense rate; allow clarification and per-step decomposition. \\
\bottomrule
\end{tabular}
\caption{Threats and intent-layer responses.}
\label{tab:threats}
\end{table}

\subsection{Out of Scope}

The threat model assumes correct static policy, complete mediation, trusted-message separation, an uncompromised gateway, and conservative adapter effect contracts. It excludes malicious server operators, dishonest domain adapters, model alignment, direct database compromise, gateway private-key extraction, unrelated side channels, and attacks outside tool authorization.

\section{System Model}

\subsection{Logical Components}

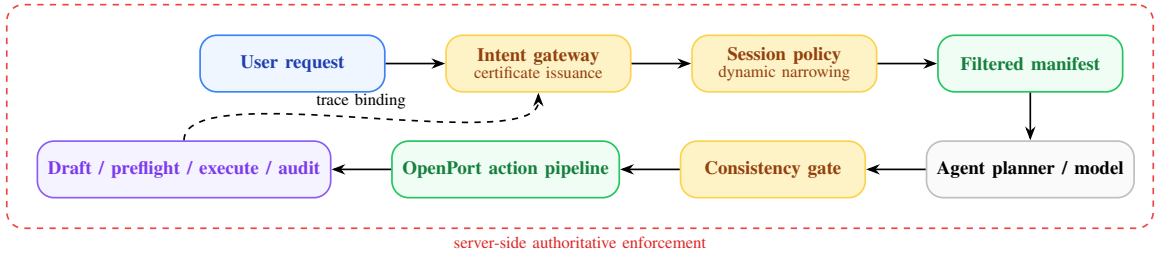
\begin{figure}[t]
\centering
\begin{floatbox}
\resizebox{\linewidth}{!}{%
\begin{tikzpicture}[
  >=Stealth,
  box/.style={draw, rounded corners=2.8mm, thick, align=center, minimum width=3.1cm, minimum height=0.95cm, inner sep=5pt, font=\small},
  edge/.style={->, thick},
  trust/.style={draw=opRed, dashed, rounded corners=2.8mm, thick, inner sep=5mm}
]
\node[box, draw=opBlue, fill=opBlueFill, text=opBlueText] (user) {\textbf{User request}};
\node[box, draw=opAmber, fill=opAmberFill, text=opAmberText, right=10mm of user] (intent) {\textbf{Intent gateway}\\[-2pt]\footnotesize certificate issuance};
\node[box, draw=opAmber, fill=opAmberFill, text=opAmberText, right=10mm of intent] (policy) {\textbf{Session policy}\\[-2pt]\footnotesize dynamic narrowing};
\node[box, draw=opGreen, fill=opGreenFill, text=opGreenText, right=10mm of policy] (manifest) {\textbf{Filtered manifest}};
\node[box, draw=black!25, fill=black!2, below=8mm of manifest] (planner) {\textbf{Agent planner / model}};
\node[box, draw=opAmber, fill=opAmberFill, text=opAmberText, left=10mm of planner] (gate) {\textbf{Consistency gate}};
\node[box, draw=opGreen, fill=opGreenFill, text=opGreenText, left=10mm of gate] (openport) {\textbf{OpenPort action pipeline}};
\node[box, draw=opPurple, fill=opPurpleFill, text=opPurpleText, left=10mm of openport] (audit) {\textbf{Draft / preflight / execute / audit}};
\draw[edge] (user) -- (intent);
\draw[edge] (intent) -- (policy);
\draw[edge] (policy) -- (manifest);
\draw[edge] (manifest) -- (planner);
\draw[edge] (planner) -- (gate);
\draw[edge] (gate) -- (openport);
\draw[edge] (openport) -- (audit);
\draw[edge, dashed] (audit.north) .. controls +(0,1.1) and +(0,-1.1) .. node[above, font=\footnotesize] {trace binding} (intent.south);
\node[trust, fit=(intent) (policy) (manifest) (gate) (openport) (audit), label={[font=\footnotesize, text=opRedText]below:server-side authoritative enforcement}] {};
\end{tikzpicture}%
}
\end{floatbox}
\caption{The intent layer over the OpenPort effect-control substrate.}
\label{fig:architecture}
\end{figure}

Figure~\ref{fig:architecture} shows the architecture. The intent-certificate layer is inserted before OpenPort's existing action pipeline.

The components are:

\begin{itemize}
  \item \textbf{User request}: a natural-language instruction, possibly with attachments or context.
  \item \textbf{Intent gateway}: derives an intent certificate from the trusted user request and request metadata.
  \item \textbf{Session policy}: a dynamic policy lease that can only narrow static integration policy.
  \item \textbf{OpenPort manifest}: a tool list filtered by static authorization and session policy.
  \item \textbf{Agent planner}: an untrusted model/planner that proposes tool calls.
  \item \textbf{Consistency gate}: checks that proposed tool, payload, and expected effects match the certificate.
  \item \textbf{OpenPort action pipeline}: existing effect-control layer for drafts, preflight, idempotency, state witnesses, and audit.
\end{itemize}

\subsection{Trust and Authority}

The gateway is authoritative. The classifier may be wrong. The planner may be wrong. The model may be manipulated. Tool metadata may be malicious. The intent layer therefore has two central trust rules:

\begin{enumerate}
  \item An intent certificate may only reduce authority.
  \item A proposed tool call must still pass static authorization and OpenPort effect-control checks.
\end{enumerate}

These rules allow the authorization layer to use probabilistic components without making them trusted roots. A high-confidence certificate can expose a narrower useful tool set. A low-confidence certificate cannot grant new scopes; it can only force clarification, draft review, or denial.

\paragraph{Static authority is not user intent}
Monotonicity over static policy is necessary but insufficient for request-level authorization. A broad or malformed certificate can remain entirely inside an integration's static scopes while still exceeding what the user asked for. The intent-certificate layer therefore provides two different guarantees: \emph{static-policy non-expansion} follows from authoritative conjunction with OpenPort policy, whereas \emph{expressed-intent confinement} is conditional on certificate fidelity, conservative effect interpretation, and complete mediation. The runtime study measures the practical gap between these guarantees as unsafe accepted authority.

\subsection{Intent-Control Layer vs Effect-Control Layer}

OpenPort Protocol already controls effects. It answers questions such as:

\begin{itemize}
  \item Does the integration have the required scope?
  \item Is the resource collection, workspace, or query window allowed?
  \item Is this action high risk?
  \item Is preflight required and bound to the payload?
  \item Is an idempotency key required?
  \item Should the request create a draft instead of executing?
\end{itemize}

The intent-certificate layer controls request justification before effects. It asks:

\begin{itemize}
  \item What action class did the user request?
  \item What resources and bounds did the user mention?
  \item Which tools are unnecessary for this request even though statically authorized?
  \item Does the proposed payload exceed the certified request?
  \item Should ambiguity reduce authority or require review?
\end{itemize}

The two layers remain separate so that request interpretation cannot bypass effect-level authorization.

\section{Intent-Certificate Authorization Design}

\subsection{Intent Certificates}

An intent certificate is a server-issued policy assertion binding a trusted request message to an authorization-relevant interpretation. It is not a natural-language summary, a PKI identity credential, or proof of the user's latent mental intent. It is a short-lived policy input with explicit classes, bounds, confidence, lifecycle state, and provenance.

A certificate contains:

\begin{itemize}
  \item \texttt{id}: stable certificate identifier;
  \item \texttt{requestHash}: hash of canonicalized user request and selected trusted metadata;
  \item \texttt{subjectBinding}: actor, app, key, tenant/workspace, session, and trusted-message identifiers;
  \item \texttt{versionBinding}: canonicalization, static-policy, and tool-registry versions;
  \item \texttt{intentClasses}: action classes such as read, summarize, create, update, delete, export, admin, unknown;
  \item \texttt{resourceBounds}: resource collections, workspaces, query windows, record IDs, or object types named by the user;
  \item \texttt{effectBounds}: maximum effect type, item count limits, mutation scope, or allowed action family;
  \item \texttt{confidence}: calibrated score or discrete confidence band;
  \item \texttt{reviewMode}: allow, draft, preflight, confirm, deny, or clarify;
  \item \texttt{expiresAt}: time or turn boundary after which the certificate cannot authorize new calls;
  \item \texttt{issuedAt}/\texttt{revokedAt}: lifecycle state used to prevent replay after revocation;
  \item \texttt{classifierSource}: rule, model, product flow, hybrid, or human;
  \item \texttt{auditDigest}: hash used for trace correlation.
\end{itemize}

The certificate should not contain raw secrets or unnecessary private data. If a request includes sensitive content, the certificate stores bounded facts and hashes rather than full text.

The reference runtime implements request hashing, app/key/actor isolation, classes, resource/effect bounds, confidence, routing, expiry, provenance, audit correlation, and in-memory revocation. The target reference profile additionally requires tenant/workspace/session/message identifiers and canonicalization, policy, and tool-registry version bindings.

\subsection{Intent Classes}

Table~\ref{tab:intentclasses} summarizes a minimal class set.

\begin{table*}[t]
\centering
\footnotesize
\begin{tabularx}{\textwidth}{L{0.12\textwidth}YY}
\toprule
\textbf{Class} & \textbf{Typical user request} & \textbf{Allowed effect envelope} \\
\midrule
read & Show, list, inspect, find. & Bounded read tools; no mutation; no bulk export unless explicit. \\
summarize & Summarize, compare, explain. & Read and transform within bounds; no write/export/delete. \\
transform & Reformat, classify, extract fields. & Compute-only transformation; drafts for writes. \\
create & Create a record or draft. & Narrow create draft; immediate execution only under low risk and static policy. \\
update & Modify named record. & Bounded update draft; target resource must match certificate. \\
delete & Delete, remove, revoke. & High-risk draft/preflight; narrow target set; no wildcard deletes. \\
export & Download, send, CSV, bulk output. & High-risk review; row/date/resource bounds required. \\
delegate & Assign, invite, send on behalf. & Review; recipient and scope bounds required. \\
admin & Configure, grant, disable, change policy. & Strong review/step-up; hidden unless explicit. \\
unknown & Ambiguous or low-confidence request. & Minimal tools, clarification, or draft-only mode. \\
\bottomrule
\end{tabularx}
\caption{Intent classes and allowed effect envelopes.}
\label{tab:intentclasses}
\end{table*}

The taxonomy defines top-level safety semantics while allowing deployments to add domain-specific subtypes.

\subsection{Session Policy Narrowing}

Given a certificate \(C\), the gateway derives a session policy \(P_C\). This policy is ephemeral and bounded by the static integration policy \(P_A\). It includes:

\begin{itemize}
  \item allowed and denied intent classes;
  \item maximum risk tier;
  \item visible tool filters;
  \item required review mode;
  \item resource and date bounds;
  \item export limits;
  \item certificate expiry.
\end{itemize}

For example, a summary request may expose only read tools and transformation helpers. A create request may expose one create tool and require draft review. A delete request may expose the delete tool only with preflight and idempotency requirements inherited from OpenPort.

\subsection{Manifest Filtering}

In many tool-use systems, the model first sees all tools available to a credential and then chooses one. Intent-aware manifest filtering reduces this surface before model selection. The filtered manifest is:

\[
\begin{aligned}
\visible_{\mathrm{intent}}(a,u)
  =\{t \in \tool \mid{}&
  t \in \visible_{\text{\sys}}(a)\\
  &{}\land P_C(u,t)=1\}.
\end{aligned}
\]

The set \(\visible_{\text{\sys}}(a)\) is the static manifest produced by the OpenPort-style substrate under scopes and data policy. The session predicate \(P_C(u,t)\) can only remove tools. Therefore:

\[
\visible_{\mathrm{intent}}(a,u)\subseteq \visible_{\text{\sys}}(a).
\]

This property is stronger than prompting the model not to call certain tools. Hidden tools are not available for ordinary selection, and unexpected attempts can be denied by the consistency gate.

\subsection{Intent-Tool-Payload Consistency}

Manifest filtering is insufficient by itself because a visible tool can still be called with an overbroad payload. The intent-certificate layer therefore checks consistency:

\[
\consistent(C,t,x,e)=1,
\]

where \(C\) is the certificate, \(t\) is the tool, \(x\) is the payload, and \(e\) is the predicted or preflighted effect. A call is consistent only if:

\begin{enumerate}
  \item tool class is allowed by the intent class;
  \item payload resource identifiers are within certificate bounds;
  \item query windows and item count limits do not exceed request bounds;
  \item effect risk is compatible with review mode;
  \item no denied class is present.
\end{enumerate}

If consistency fails, the enforce profile denies the proposal or asks for clarification; it does not silently turn an intent-inconsistent proposal into an accepted draft. A consistent create action may still be routed to a draft by the substrate's effect controls. This distinction matters empirically: a draft is not a completed domain effect, but it is persistent accepted authority and therefore counts as unsafe acceptance when it is inconsistent with the reference request.

\subsection{Routing Outcomes}

The authorization layer returns one of six routing outcomes:

\[
\begin{aligned}
\mathcal{R}=\{&
\mcode{deny},\mcode{clarify},\mcode{draft},\\
&\mcode{preflight},\mcode{confirm},\mcode{execute}\}.
\end{aligned}
\]

These outcomes are not a total-order lattice. \mcode{deny} and \mcode{clarify} admit no authority path; \mcode{draft}, \mcode{preflight}, and \mcode{confirm} create or preserve a governed path without a completed domain effect; only \mcode{execute} permits an immediate effect. The certificate supplies a maximum execution mode, and the static substrate may route a consistent proposal more conservatively. A high-risk or low-confidence proposal therefore cannot become \mcode{execute}, but a review-routed proposal must still satisfy the request bounds before any draft or preflight artifact is accepted.

\subsection{Confidence-Aware Policy}

The intent classifier is explicitly not a trusted principal. Its output is a policy hint that may narrow authority, but it cannot create authority that static OpenPort policy withheld. The authorization layer therefore treats confidence as an input to fail-closed routing rather than as proof that an effect is safe.

Let \(\gamma\in[0,1]\) be the certificate confidence. A deployment may calibrate thresholds, but the reference policy uses three bands:

\[
b(\gamma)=
\begin{cases}
\mcode{low}, & \gamma < \gamma_L,\\
\mcode{medium}, & \gamma_L \leq \gamma < \gamma_H,\\
\mcode{high}, & \gamma \geq \gamma_H.
\end{cases}
\]

The policy decision is ordered so that hard safety constraints dominate confidence. Static denial is checked first; conflicting intent is denied or split before execution; compound workflows receive per-step certificates; low confidence triggers clarification; high-risk effects require explicit confirmation and OpenPort preflight; medium confidence uses draft/preflight; only high-confidence low-risk tasks can expose bounded tools for immediate execution. Table~\ref{tab:confidencepolicy} summarizes the policy.

\begin{table*}[t]
\centering
\footnotesize
\begin{tabularx}{\textwidth}{L{0.25\textwidth}L{0.23\textwidth}Y}
\toprule
\textbf{Condition} & \textbf{Server action} & \textbf{Security meaning} \\
\midrule
Static policy denies tool or scope & deny & Classifier cannot widen OpenPort authority. \\
Conflicting intent classes & deny or split workflow & One certificate cannot authorize incompatible effects. \\
Multi-step request & per-step certificates & No broad long-lived certificate for a compound workflow. \\
Low confidence & clarification & Uncertain intent cannot authorize broad tools. \\
High-risk intent & confirmation and preflight & Destructive/export/admin effects require explicit review binding. \\
Medium confidence & draft or preflight & Plausible but uncertain action is review-routed. \\
High confidence and low risk & allow bounded tools & Only narrow low-risk effects may execute directly. \\
\bottomrule
\end{tabularx}
\caption{Confidence-aware intent-certificate routing. The classifier may only reduce authority; low confidence, conflict, high risk, and static denial all move the decision away from immediate execution.}
\label{tab:confidencepolicy}
\end{table*}

Classifier outputs may only narrow the useful manifest. Unsafe confidence states route to clarification, draft, preflight, confirmation, splitting, or denial; a high-confidence but statically unauthorized certificate denies, and a low-confidence certificate never falls back to broad static scopes.

\subsection{Audit Binding}

Every decision should be reconstructable:

\[
\begin{aligned}
\text{request}&\rightarrow\text{certificate}\rightarrow\text{policy}
\rightarrow\text{manifest}\\
&\rightarrow\text{tool call}\rightarrow\text{decision}
\rightarrow\text{outcome}.
\end{aligned}
\]

The audit stream should include certificate identifiers and hashes, not necessarily raw user text. This supports incident response without leaking sensitive prompt content into logs.

\section{Formal Model}

\subsection{Entities}

Let:

\begin{itemize}
  \item \(U\) be the set of user requests;
  \item \(A\) be the set of integration apps;
  \item \(K\) be the set of credentials;
  \item \(R\) be the set of resources;
  \item \(\tool\) be the set of tools;
  \item \(X_t\) be the payload space for tool \(t\);
  \item \(E_t\) be the effect space for tool \(t\);
  \item \(S(a)\) be static scopes granted to app \(a\);
  \item \(P_A\) be static OpenPort policy;
  \item \(C(u)\) be an intent certificate derived from request \(u\);
  \item \(P_C\) be session policy derived from \(C(u)\).
\end{itemize}

Each tool \(t\) has metadata:

\[
\begin{aligned}
M(t)=(&\mathrm{ReqScopes}(t),\mathrm{Risk}(t),\\
&\mathrm{Kind}(t),\mathrm{Schema}(t)).
\end{aligned}
\]

\subsection{Static OpenPort Authorization}

The reference substrate defines a visible static tool set:

\[
\begin{aligned}
\visible_{\sys}(a)
  =\{t\in\tool \mid{}&
  \mathrm{ReqScopes}(t)\subseteq S(a)\\
  &{}\land P_A(a,t)=1\}.
\end{aligned}
\]

For a concrete call \((t,x)\), static authorization is:

\[
\begin{aligned}
\allow_{\sys}(a,t,x)={}&
\mathrm{Authn}(a)\land\mathrm{ScopeOk}(a,t)\\
&{}\land\mathrm{PolicyOk}(a,t,x)\\
&{}\land\mathrm{RiskGate}_{\sys}(a,t,x).
\end{aligned}
\]

The risk gate includes draft-first behavior, preflight, idempotency, and state-witness checks where configured.

\subsection{Intent Certificate}

An intent certificate is:

\[
\begin{aligned}
C=(&h_u,b,I_C^+,I_C^-,B_R^C,B_E^C,\\
&\gamma,\rho,\tau,\sigma,v).
\end{aligned}
\]

where \(h_u\) is the request hash; \(b\) binds the actor, app, key, tenant/workspace, session, and trusted message; \(I_C^+\) and \(I_C^-\) are allowed and denied intent classes; \(B_R^C\) and \(B_E^C\) are resource- and effect-bound predicates; \(\gamma\in[0,1]\) is confidence; \(\rho\in\mathcal{R}\) is the maximum routing outcome; \(\tau\) is expiry; \(\sigma\) identifies classifier provenance; and \(v\) binds canonicalization, static-policy, and tool-registry versions.

Validity is:

\[
\begin{aligned}
\mathrm{Valid}(C,u,t_{\mathrm{now}})={}&
(h_u=H(u))\land\mathrm{BindOk}(b)\\
&{}\land\mathrm{VersionOk}(v)
\land(t_{\mathrm{now}}<\tau)\\
&{}\land\neg\mathrm{Revoked}(C).
\end{aligned}
\]

Confidence does not make a certificate authentic or valid. It only constrains routing: a low-confidence certificate may be retained for audit or clarification, but it cannot authorize \mcode{execute}.

\subsection{Reference-Envelope and Effect Soundness}

Let an independently specified reference envelope for the trusted expressed request be

\[
G(u)=(I_u^+,I_u^-,B_R^u,B_E^u).
\]

\(G(u)\) is an independently specified operational authorization reference for the expressed request. A certificate is a field-wise refinement of \(G(u)\) when

\[
\begin{aligned}
\mathrm{CertRefines}(C,G(u))\equiv{}&
 I_C^+\subseteq I_u^+\land I_C^-\supseteq I_u^-\\
&{}\land \forall x:\ B_R^C(x)\Rightarrow B_R^u(x)\\
&{}\land \forall e:\ B_E^C(e)\Rightarrow B_E^u(e).
\end{aligned}
\]

Let \(\widehat e=\mathrm{Eff}(t,x)\) be the conservative effect bound computed from schema, tool metadata, adapter contracts, or preflight. Effect interpretation is sound when the realized domain effect refines that bound:

\[
\mathrm{EffectSound}(t,x)\equiv e_{\mathrm{real}}(t,x)\preceq \widehat e.
\]

Effect predicates are required to be downward closed:
\(B_E(e)=1\land e'\preceq e\Rightarrow B_E(e')=1\).
Neither field-wise refinement nor effect soundness follows from classifier confidence. They are explicit, independently testable contracts for certificate generators and tool adapters. When either cannot be established, the conservative routing rule forbids immediate execution.

\subsection{Monotone Session Policy and Outcomes}

The session policy \(P_C\) is monotone if:

\[
\forall a,u:\ \visible_{\mathrm{intent}}(a,u)\subseteq \visible_{\sys}(a).
\]

For a proposed call, define eligibility using the conservatively estimated effect:

\begin{equation}
\label{eq:intent-eligible}
\begin{aligned}
\widehat e&=\mathrm{Eff}(t,x),\\
\mathrm{Eligible}_{\mathrm{intent}}(a,u,t,x)
&=\allow_{\sys}(a,t,x)\\
&\quad{}\land\mathrm{Valid}(C(u),u,t_{\mathrm{now}})\\
&\quad{}\land\consistent(C(u),t,x,\widehat e).
\end{aligned}
\end{equation}

The gateway returns \(r=\mathrm{Route}_{\mathrm{intent}}(a,u,t,x)\in\mathcal{R}\). A missing, invalid, expired, or inconsistent certificate in the enforce profile maps only to \mcode{deny} or \mcode{clarify}; it cannot silently fall back to static authorization. For an eligible call, certificate constraints and substrate risk controls are merged conservatively. We distinguish an accepted authority path from an executed effect:

\[
\begin{aligned}
\allow_{\mathrm{intent}}^{\mathrm{path}}=1
\Longleftrightarrow{}&
\mathrm{Eligible}_{\mathrm{intent}}=1\\
&{}\land r\in\{\mcode{draft},\mcode{preflight},\\
&\hspace{3.5em}\mcode{confirm},\mcode{execute}\}.
\end{aligned}
\]

\[
\allow_{\mathrm{intent}}^{\mathrm{exec}}=1
\Longleftrightarrow
\mathrm{Eligible}_{\mathrm{intent}}=1\land r=\mcode{execute}.
\]

Thus a draft can be an unsafe accepted-authority event without being an unsafe executed effect. This hierarchy is used by the evaluation metrics.

\subsection{Consistency Predicate}

Let \(Class(t)\) map a tool to effect classes and \(\widehat e=\mathrm{Eff}(t,x)\). Then:

\[
\begin{aligned}
\consistent(C,t,x,\widehat e) ={}&
Class(t)\subseteq I_C^+ \land Class(t)\cap I_C^-=\emptyset\\
&\land B_R^C(x)=1 \land B_E^C(\widehat e)=1.
\end{aligned}
\]

Routing then enforces:

\[
\begin{aligned}
&\risk(t,x,\widehat e)>\theta_C
\lor\gamma<\gamma_{\min}\\
&\quad{}\lor\neg\mathrm{Boundable}(C,t,x,\widehat e)
\Rightarrow r\neq\mcode{execute}.
\end{aligned}
\]

In addition, \(\neg\consistent(C,t,x,\widehat e)\Rightarrow
r\in\{\mcode{deny},\mcode{clarify}\}\) in the enforce profile.

\subsection{Least-Privilege Objective}

For a request \(u\), utility \(Utility(u,P_C)\), and minimum utility threshold \(\theta\), the ideal session policy minimizes exposed effects:

\[
\min_{P_C} |\effects(P_C)|
\quad \text{s.t.}\quad
Utility(u,P_C)\geq \theta \land P_C\preceq P_A.
\]

The system approximates this objective through a finite intent taxonomy, static tool metadata, resource bounds, and review-mode rules. The objective is not solved by the LLM. It is encoded in gateway policy.

\subsection{Intent-Trace Field Completeness}

Define a trace \(T_i\) as field-complete if it contains:

\[
(h_u, C, P_C, \visible, t, x, decision, reason, outcome).
\]

Intent-trace field completeness over \(N\) traces is:

\[
TFC=\frac{1}{N}\sum_{i=1}^{N}\mathbf{1}[T_i\ \text{is field-complete}].
\]

This metric measures field presence; it does not assess human interpretation of the trace.

\subsection{Security Propositions and Proof Sketches}

The propositions assume complete mediation by the gateway. Propositions~1--3 require only the construction above; the request-confinement contract additionally requires an independently specified reference envelope, certificate refinement, and effect soundness.

\statementheading{Proposition 1 (Manifest Monotonicity)}
For any authenticated actor \(a\) and request \(u\),
\[
\visible_{\mathrm{intent}}(a,u)\subseteq \visible_{\sys}(a).
\]
\emph{Proof sketch.} By definition, the intent-filtered manifest is derived by filtering the statically visible OpenPort manifest through certificate classes, bounds, and review policy. No construction in \(C(u)\) adds tools that were not already in \(\visible_{\sys}(a)\). Therefore classifier output, prompt injection, or certificate corruption may suppress tools, but cannot reveal a tool outside static OpenPort authorization.

\statementheading{Proposition 2 (Static-Policy Non-Expansion)}
For any actor \(a\), request \(u\), tool \(t\), and payload \(x\),
\[
\allow_{\mathrm{intent}}^{\mathrm{path}}(a,u,t,x)=1
\Rightarrow \allow_{\sys}(a,t,x)=1.
\]
\emph{Proof sketch.} Path acceptance requires eligibility, and Eq.~\eqref{eq:intent-eligible} includes \(\allow_{\sys}\) as a conjunct. Hence an over-broad or compromised classifier cannot transform a statically denied call into an accepted path, although it may fail to narrow authority to the user's expressed request.

\statementheading{Proposition 3 (High-Risk Fail-Closed Routing)}
If any of the following holds,
\[
\begin{aligned}
\risk(t,x,\widehat e)&>\theta_C,\qquad
\gamma<\gamma_{\min},\\
\neg\mathrm{Boundable}(C,t,x,\widehat e)&.
\end{aligned}
\]
then the intent-certificate layer must not authorize immediate execution:
\[
\allow_{\mathrm{intent}}^{\text{exec}}(a,u,t,x)=0.
\]
\emph{Proof sketch.} The routing constraint excludes \mcode{execute} for high-risk, low-confidence, or unboundable proposals. A consistent proposal may still enter a governed draft/preflight/confirmation path; an inconsistent proposal must deny or clarify.

\statementheading{Proposition 4 (Compositional Request-Confinement Contract)}
Assume complete mediation, an independently specified \(G(u)\), \(\mathrm{CertRefines}(C(u),G(u))\), and \(\mathrm{EffectSound}(t,x)\). Then
\[
\allow_{\mathrm{intent}}^{\mathrm{exec}}(a,u,t,x)=1\Rightarrow
\begin{cases}
Class(t)\subseteq I_u^+,\\
Class(t)\cap I_u^-=\emptyset,\\
B_R^u(x)=1,\\
B_E^u(e_{\mathrm{real}})=1.
\end{cases}
\]
\emph{Proof sketch.} Execution implies certificate consistency. Field-wise refinement maps each accepted certificate field into the independently specified request envelope; effect soundness maps the realized effect into the conservative estimate. Classifier correctness remains an independent requirement. If either contract fails, only static-policy non-expansion remains guaranteed. The nonzero unsafe-accepted-authority cases in Section~\ref{sec:evaluation} empirically illustrate this distinction.

\section{OpenPort Implementation Mapping}

\subsection{Reference Substrate}

The OpenPort-style governance substrate assumed by the intent-certificate layer provides the lower layer needed for intent enforcement. The reference architecture's agent engine exposes a manifest filtered by the tool registry. The registry declares required scopes and risk tiers. Policy helpers enforce scopes, resource allowlists, bounded query windows, and tenant/workspace boundaries. The action pipeline computes preflight impact, stores preflight records, creates drafts, handles high-risk auto-execute requirements, validates idempotency, records policy snapshots, checks state witnesses, executes drafts, and emits audit events. A conformant substrate should cover high-risk preflight and idempotency, state witness validation, stable hashes, draft approval, and security controls.

These properties support the following intent-layer assumptions:

\begin{itemize}
  \item the server owns the authoritative tool manifest;
  \item tools carry governance metadata including scopes and risk tiers;
  \item writes pass through draft/preflight execution;
  \item audit is a first-class control;
  \item static scopes and policy remain enforceable regardless of model behavior.
\end{itemize}

\subsection{Reference Intent-Layer Runtime}

The reference implementation adds a reusable intent-layer path on top of the OpenPort-style substrate and exercises real HTTP endpoints and the substrate action pipeline.

\begin{itemize}
  \item \texttt{POST /api/agent/v1/intent} creates an intent certificate with request hash, classes, resource/effect bounds, confidence, review mode, classifier source, expiry, and audit digest.
  \item \texttt{GET /api/agent/v1/manifest} loads the \texttt{intentCertificateId} bound to the authenticated app/key/user and filters the manifest after static OpenPort scope checks.
  \item \texttt{POST /api/agent/v1/preflight} accepts \texttt{intentCertificateId} and rejects intent-tool or payload-bound mismatches before producing an impact hash.
  \item \texttt{POST /api/agent/v1/actions} accepts \texttt{intentCertificateId}, checks intent-tool-payload consistency, routes high-risk execution to review, records certificate metadata in draft policy snapshots, and emits audit events with certificate correlation.
  \item The in-memory store contains an intent certificate table scoped by app, key, actor, and expiry. A certificate created by one key cannot be reused by another key.
\end{itemize}

The implementation is concentrated in the intent engine, store, agent engine, runtime, and endpoint wiring modules. Endpoint tests cover monotonic manifest narrowing, reason-code denials, payload-bound failure, high-risk review routing, audit binding, draft policy snapshots, and cross-key certificate isolation.

\begin{table*}[t]
\centering
\footnotesize
\begin{tabularx}{\textwidth}{L{0.25\textwidth}Y}
\toprule
\textbf{Test group} & \textbf{Covered invariant} \\
\midrule
Monotonic manifest narrowing & \(\visible_{\mathrm{intent}}\subseteq\visible_{\sys}\). \\
Intent reason codes & Denials return stable \texttt{agent.intent\_*} codes. \\
Payload-bound failure & Out-of-bound resources fail with \path{agent.intent_payload_exceeds_bound}. \\
High-risk review routing & High-risk effects cannot immediately execute through an intent certificate. \\
Audit binding & Certificate metadata appears in audit events and draft policy snapshots. \\
Cross-key isolation & A certificate created by one key is not reusable by another key. \\
\bottomrule
\end{tabularx}
\caption{Endpoint-test summary for the reference intent-layer runtime.}
\label{tab:endpointtests}
\end{table*}

\subsection{Reference Hardening and Deployment Profile}

The reference implementation evaluates the following hardening controls:

\begin{table}[t]
\centering
\small
\resizebox{\linewidth}{!}{%
\begin{tabular}{L{0.34\linewidth}L{0.44\linewidth}}
\toprule
\textbf{Reference-hardening item} & \textbf{Evaluated evidence} \\
\midrule
Certificate revocation and ephemeral-state GC & Tests cover in-memory revocation and sweep behavior. \\
Tamper-evident local audit sequence & Tests verify stable previous/current audit hashes. \\
Intent endpoint and action invariants & Endpoint tests verify manifest narrowing, stable reason codes, audit binding, and cross-key isolation. \\
OpenPort security conformance profile & Security conformance tests cover state witness, idempotency replay, rate limiting, and audit completeness. \\
Adversarial and malformed-input coverage & Adversarial and fuzz tests preserve stable envelopes and reject malformed inputs. \\
Rate-limit and abuse controls & Abuse tests verify denial without draft/execution side effects. \\
\bottomrule
\end{tabular}
}
\caption{Reference-hardening controls exercised by the intent-layer prototype.}
\label{tab:referencehardening}
\end{table}

The production target additionally requires:

\begin{enumerate}
  \item persistent certificate storage with crash-safe revocation and retention semantics;
  \item calibrated classifier implementations beyond rule/provided certificates;
  \item richer per-tool consistency predicates for complex workflows;
  \item certificate expiry by turn count or completed effect, not only wall-clock time;
  \item wider malformed-certificate fuzzing and adversarial certificate tests;
  \item signed or externally anchored audit events beyond the local hash chain;
  \item distributed rate limiting plus concurrency/replay tests beyond the process-local reference runtime;
  \item policy versioning and adapter effect-contract tests for tool semantics;
  \item a release-quality conformance profile and a production profile for intent-aware governance.
\end{enumerate}

Two target profiles distinguish reusable semantics from operational hardening. The \textbf{reference profile} requires app/key/actor-scoped certificate isolation, intent-aware manifest filtering, intent-bound preflight/action checks, stable reason codes, audit binding, monotonicity over static OpenPort policy, and local lifecycle cleanup. The prototype implements these semantics in an in-memory runtime. The \textbf{production profile} additionally requires persistent certificate state, externally verifiable audit integrity, distributed replay and rate-limit behavior, policy-version binding, and adapter effect contracts.

\subsection{Backward Compatibility}

Backward compatibility is provided by endpoint and deployment-mode separation. A legacy or \mcode{off} endpoint retains static OpenPort semantics. An intent-aware endpoint in \mcode{enforce} or \mcode{conformance} mode requires a valid certificate: a missing, malformed, expired, revoked, cross-subject, or version-stale certificate yields a minimal clarification manifest or denial. A supplied malformed certificate does not restore broad static behavior. \mcode{shadow} mode records counterfactual decisions without enforcement and marks its responses as non-enforcing.

\subsection{Reason Codes}

The existing OpenPort reason-code style should be extended with:

\begin{itemize}
  \item \texttt{agent.intent\_not\_found}
  \item \texttt{agent.intent\_invalid}
  \item \texttt{agent.intent\_binding\_mismatch}
  \item \texttt{agent.intent\_version\_stale}
  \item \texttt{agent.intent\_revoked}
  \item \texttt{agent.intent\_low\_confidence}
  \item \texttt{agent.intent\_conflicting}
  \item \texttt{agent.intent\_tool\_mismatch}
  \item \texttt{agent.intent\_payload\_exceeds\_bound}
  \item \texttt{agent.intent\_expired}
  \item \texttt{agent.intent\_review\_required}
\end{itemize}

These codes allow deterministic client recovery. For example, \texttt{agent.intent\_low\_confidence} can ask the client to clarify, while \texttt{agent.intent\_payload\_exceeds\_bound} should not be retried unchanged. The reference implementation emits core not-found, mismatch, bound, expiry, and review codes; the target reference profile also distinguishes binding, version, and revocation failures.

\section{Protocol Extension Sketch}

The intent layer can be exposed as a conservative OpenPort extension while preserving compatibility with existing clients.

\subsection{Intent-Aware Manifest Request}

An intent-aware client may request a narrowed manifest by sending a certificate reference or a request body to a server-side intent endpoint:

\begin{lstlisting}[language=json,basicstyle=\ttfamily\footnotesize]
POST /api/agent/v1/intent
{
  "request": "Summarize recent records for this period",
  "context": {
    "trustedUserMessageId": "msg_123",
    "untrustedContextPresent": true
  }
}
\end{lstlisting}

The server returns a certificate:

\begin{lstlisting}[language=json,basicstyle=\ttfamily\footnotesize]
{
  "intentCertificateId": "intent_123",
  "requestHash": "sha256:...",
  "intentClasses": ["read", "summarize"],
  "reviewMode": "allow",
  "confidence": 0.87,
  "expiresAt": "2026-06-17T23:59:59Z"
}
\end{lstlisting}

The client then calls:

\begin{lstlisting}[basicstyle=\ttfamily\footnotesize]
GET /api/agent/v1/manifest?intentCertificateId=intent_123
\end{lstlisting}

An implementation can also embed the certificate in an authenticated request header. The important property is that the server validates the certificate and computes filtering. The client must not be trusted to filter tools locally.

\subsection{Action Request Extension}

The action endpoint accepts an optional certificate reference:

\begin{lstlisting}[language=json,basicstyle=\ttfamily\footnotesize]
POST /api/agent/v1/actions
{
  "action": "resource.create",
  "payload": { "...": "..." },
  "intentCertificateId": "intent_123",
  "execute": false
}
\end{lstlisting}

The server performs three checks:

\begin{enumerate}
  \item static OpenPort authorization;
  \item certificate validity and monotone session policy;
  \item intent-tool-payload consistency.
\end{enumerate}

Only after those checks does the existing OpenPort action pipeline create a draft, require preflight, or execute under configured policy.

\subsection{Algorithm 1: Manifest Narrowing}

\begin{lstlisting}[basicstyle=\ttfamily\footnotesize]
function intentAwareManifest(ctx, certificateId):
    staticTools = baseManifest(ctx)
    if deploymentMode(ctx) is legacy or off:
        return staticTools
    cert = validateBoundCertificate(ctx, certificateId)
    if cert is missing:
        audit("agent.intent.missing")
        return minimalClarificationTools()
    if cert is invalid or expired:
        audit("agent.intent.invalid")
        return minimalClarificationTools()
    sessionPolicy = deriveSessionPolicy(cert)
    visible = []
    for tool in staticTools:
        if sessionPolicy.allowsTool(tool):
            visible.append(tool)
    audit("agent.intent.manifest", cert.id, count(visible))
    return visible
\end{lstlisting}

The call to \texttt{baseManifest(ctx)} must happen before session filtering. This ordering enforces monotonicity. A session policy never sees tools that static policy has already hidden.

\subsection{Algorithm 2: Action Consistency}

\begin{lstlisting}[basicstyle=\ttfamily\footnotesize]
function createGovernedAction(ctx, input):
    tool = resolveTool(input.action)
    requireStaticAuth(ctx, tool, input.payload)

    if deploymentMode(ctx) is legacy or off:
        return governedAction(ctx, input)
    cert = validateBoundCertificate(ctx,
        input.intentCertificateId)
    if cert is missing or invalid or expired:
        return deny("agent.intent_invalid")

    effect = estimateOrPreflightEffect(ctx, tool, input.payload)
    if not consistent(cert, tool, input.payload, effect):
        return deny("agent.intent_tool_mismatch")

    riskMode = substrateRiskMode(tool, input)
    mode = mergeReviewModes(cert.reviewMode, riskMode)
    return governedAction(ctx,
        applyMode(input, mode))
\end{lstlisting}

The design deliberately reuses the substrate's existing write pipeline. The intent-certificate layer should not duplicate draft, preflight, idempotency, or execution logic.

\subsection{Algorithm 3: Certificate Expiry and Drift}

Long-running conversations create drift risk. A certificate should expire after a time interval, turn count, context transition, or completed effect. The server can define:

\[
\begin{aligned}
\mathrm{Expired}(C)={}&
t_{\mathrm{now}}>\tau_C
\lor \mathrm{turns}(C)>n_C\\
&{}\lor \mathrm{effectCompleted}(C)=1.
\end{aligned}
\]

If a certificate expires, the gateway can reclassify the latest trusted user request, ask for clarification, or force draft review. For multi-step plans, per-step certificates are preferable to a single broad certificate.

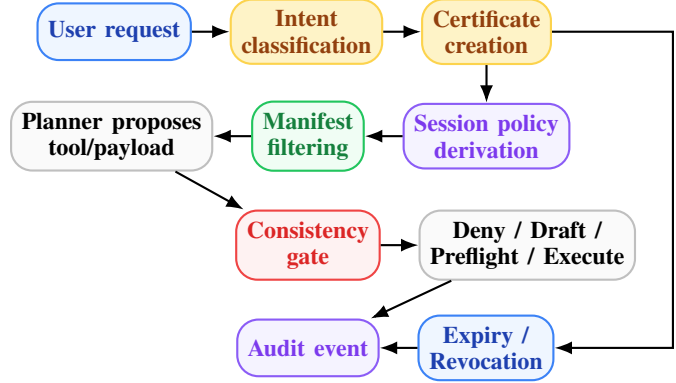
\begin{figure*}[t]
\centering
\begin{floatbox}
\begin{tikzpicture}[
  node distance=5mm and 5mm,
  box/.style={draw, rounded corners=2.8mm, thick, align=center, minimum height=7.5mm, inner sep=4pt, font=\small},
  arrow/.style={-Latex, thick}
]
\node[box, draw=opBlue, fill=opBlueFill, text=opBlueText] (u) {\textbf{User request}};
\node[box, draw=opAmber, fill=opAmberFill, text=opAmberText, right=of u] (c1) {\textbf{Intent}\\\textbf{classification}};
\node[box, draw=opAmber, fill=opAmberFill, text=opAmberText, right=of c1] (c2) {\textbf{Certificate}\\\textbf{creation}};
\node[box, draw=opPurple, fill=opPurpleFill, text=opPurpleText, below=of c2] (p) {\textbf{Session policy}\\\textbf{derivation}};
\node[box, draw=opGreen, fill=opGreenFill, text=opGreenText, left=of p] (m) {\textbf{Manifest}\\\textbf{filtering}};
\node[box, draw=black!25, fill=black!2, left=of m] (pl) {\textbf{Planner proposes}\\\textbf{tool/payload}};
\node[box, draw=opRed, fill=opRedFill, text=opRedText, below=of m] (g) {\textbf{Consistency}\\\textbf{gate}};
\node[box, draw=black!25, fill=black!2, right=of g] (o) {\textbf{Deny / Draft /}\\\textbf{Preflight / Execute}};
\node[box, draw=opPurple, fill=opPurpleFill, text=opPurpleText, below=of g] (a) {\textbf{Audit event}};
\node[box, draw=opBlue, fill=opBlueFill, text=opBlueText, right=of a] (x) {\textbf{Expiry /}\\\textbf{Revocation}};

\draw[arrow] (u) -- (c1);
\draw[arrow] (c1) -- (c2);
\draw[arrow] (c2) -- (p);
\draw[arrow] (p) -- (m);
\draw[arrow] (m) -- (pl);
\draw[arrow] (pl) -- (g);
\draw[arrow] (g) -- (o);
\draw[arrow] (o) -- (a);
\draw[arrow] (c2.east) -- ++(16mm,0) |- ([xshift=8mm]x.east) -- (x.east);
\draw[arrow] (x) -- (a);
\end{tikzpicture}
\end{floatbox}
\caption{Operational lifecycle of an intent certificate. The security boundary spans the server-side path from certificate creation through manifest narrowing, consistency gating, outcome routing, audit, and certificate expiry/revocation.}
\label{fig:certlifecycle}
\end{figure*}

\subsection{Deployment Modes}

The intent layer can be rolled out in stages:

\begin{itemize}
  \item \textbf{off}: existing OpenPort behavior;
  \item \textbf{shadow}: compute certificates and audit decisions, but do not enforce;
  \item \textbf{compatibility draft-only}: low-confidence or legacy mismatches may become drafts and must be counted as accepted authority;
  \item \textbf{enforce}: hide/deny/draft according to policy;
  \item \textbf{conformance}: fail tests if monotonicity or reason-code invariants break.
\end{itemize}

Shadow mode supports calibration. Security enforcement and its evaluation begin in enforce mode.

\section{Evaluation}
\label{sec:evaluation}

The evaluation combines deterministic conformance checks, endpoint tests, a runtime-backed synthetic microbenchmark, a real-model classifier/planner study, a scored end-to-end runtime study, and a benchmark-family-inspired transfer subset. Together they test static-policy non-expansion, request-envelope enforcement, model-induced failure modes, and the distinction between accepted authority and completed effects.

Table~\ref{tab:evidencetypes} separates deterministic conformance checks, trace analysis, and API-backed runtime trials by evidence type.

\begin{table}[t]
\centering
\small
\resizebox{\linewidth}{!}{%
\begin{tabular}{L{0.24\linewidth}L{0.18\linewidth}L{0.15\linewidth}L{0.31\linewidth}}
\toprule
\textbf{Evidence type} & \textbf{Exercises runtime?} & \textbf{Uses LLM?} & \textbf{Supports} \\
\midrule
Endpoint tests & Yes & No & Protocol/runtime invariants and regression coverage. \\
Synthetic sanity scripts & No & No & Policy logic, ablation intuition, and confidence routing checks. \\
Runtime-backed microbenchmark & Yes & No & HTTP-path checks over 176 deterministic instances generated from author-designed templates. \\
Benchmark-family transfer set & Yes & No & Runtime-backed scoring of 25 locally authored tasks preserving structures from four benchmark families. \\
Expanded real-LLM pilot & No runtime action path & Yes & API-backed classifier/planner behavior outside the runtime action path. \\
Small end-to-end LLM runtime benchmark & Yes & Yes & API-backed runtime-path evidence over scored synthetic tasks. \\
Transfer-subset end-to-end LLM runtime pilot & Yes & Yes & API-backed runtime-path evidence over 12 benchmark-family-inspired tasks. \\
Trace-backed normalizer ablation & Existing runtime traces only & No new calls & Counterfactual certificate-bound analysis over frozen traces. \\
\bottomrule
\end{tabular}
}
\caption{Evidence types in the intent-layer evaluation.}
\label{tab:evidencetypes}
\end{table}

\subsection{Research Questions}

\begin{itemize}
  \item RQ1: How does the intent layer change unjustified tool exposure and accepted-authority paths under controlled deterministic comparisons?
  \item RQ2: What accepted-authority and completed-effect outcomes occur when real-model certificates and plans traverse the combined intent-layer and OpenPort path?
  \item RQ3: What completion, clarification, review, and latency costs accompany the observed containment?
  \item RQ4: In which certificate fields and effect classes do residual unsafe accepted-authority cases concentrate?
  \item RQ5: What roles do manifest filtering, payload consistency, confidence routing, and trace-backed certificate normalization play?
\end{itemize}

\subsection{Baselines}

\begin{table}[t]
\centering
\footnotesize
\begin{tabularx}{\linewidth}{L{0.10\linewidth}YL{0.23\linewidth}}
\toprule
\textbf{Cond.} & \textbf{Description} & \textbf{Where evaluated} \\
\midrule
B0 & Model-only tool calling with a static list and no gateway. & Toy check only. \\
B1 & OpenPort static policy and effect controls without intent certificates. & Toy and 176-instance runtime check. \\
B2 & OpenPort plus intent certificates, manifest filtering, and consistency controls. & All runtime studies. \\
B3 & Prompt-only instruction without server-side intent enforcement. & Toy ablation only. \\
\bottomrule
\end{tabularx}
\caption{Evaluated conditions and their scope.}
\label{tab:baselines}
\end{table}

\subsection{Task Suites}

The evaluation uses four synthetic task suites:

\begin{itemize}
  \item a 12-task deterministic policy sanity set;
  \item 176 deterministic runtime instances generated from author-designed benign, mismatch, injection, poisoning, ambiguity, and compound-workflow templates;
  \item a 34-task parameterized runtime set repeated three times per model over three synthetic resource collections;
  \item 25 locally authored transfer tasks preserving structural patterns from AgentDojo, ToolSandbox, tau-bench, and ToolEmu, including a 12-task real-model runtime subset.
\end{itemize}

All tasks and fixtures use synthetic data. The task authors specified reference certificates, allowed/denied tools, resource bounds, and effect bounds directly in the fixtures. These are \emph{author-specified reference certificates}; the study did not include independent annotation.

The 25-task transfer set contains six AgentDojo-shaped prompt-injection tasks, seven ToolSandbox-shaped tasks spanning canonicalization, insufficient information, and state dependency, six tau-bench-shaped policy tasks, and six ToolEmu-shaped risk cases. These are locally authored structural analogues; the original task instances and simulators are not used.

\begin{table*}[t]
\centering
\footnotesize
\begin{tabularx}{\textwidth}{L{0.30\textwidth}L{0.18\textwidth}Y}
\toprule
\textbf{Source family} & \textbf{Construction} & \textbf{Role in evaluation} \\
\midrule
OpenPort synthetic runtime suite & Author-designed synthetic tasks & Runtime-backed microbenchmark and scored end-to-end runtime study. \\
AgentDojo / InjecAgent & Locally authored prompt-injection analogues & Prompt-injection transfer cases. \\
ToolHijacker / ToolTweak / MCPTox & Locally authored tool-poisoning analogues & Tool-metadata attack transfer cases. \\
AgentDojo / ToolSandbox / tau-bench / ToolEmu & Twenty-five locally authored structural analogues & Runtime-backed transfer evaluation. \\
\bottomrule
\end{tabularx}
\caption{Task-source scope. Benchmark-family transfer is distinct from original-benchmark reproduction.}
\label{tab:benchmarkscope}
\end{table*}

\subsection{Metrics}

\paragraph{Unsafe Tool Exposure Rate}

\[
UTER=\frac{\#\{\text{unjustified high-risk tools visible}\}}{\#\{\text{unjustified high-risk tools}\}}.
\]

\paragraph{Unsafe Accepted-Authority Rate}

\[
UAR_{\mathrm{path}}=
\frac{\#\left\{\substack{\text{intent-inconsistent draft, preflight,}\\
\text{confirmation, or execution paths accepted}}\right\}}
{\#\{\text{attack or mismatch tasks}\}}.
\]

An accepted draft counts because it creates persistent reviewable authority even when no domain effect completes. The deterministic microbenchmark predates this disaggregation and reports a conservative composite indicator \(UAR_c\) that is one when either unjustified high-risk exposure or an inconsistent action path occurs. The real-model runtime studies report path acceptance directly. These UAR quantities are reported separately as distinct estimands.

\paragraph{Unsafe Execution Rate}

\[
UER=\frac{\#\{\text{unsafe successful executions}\}}{\#\{\text{attack or mismatch tasks}\}}.
\]

\paragraph{Strict Completion and Governed Artifact Rate}

\[
BCR_{\text{strict}}=
\frac{\#\left\{\substack{\text{benign tasks completed}\\
\text{without a review artifact}}\right\}}
{\#\{\text{benign tasks}\}}.
\]

\[
GAR=
\frac{N_{\text{strict benign}}+
N_{\text{reviewable benign artifact}}}
{N_{\text{benign}}}.
\]

This distinction matters because a draft or preflight artifact may be a correct governance outcome without completing the user's task. Earlier artifact fields named ``safe BCR'' map to \(GAR\), while strict task completion is reported separately.

\begin{table}[t]
\centering
\small
\begin{tabular}{L{0.36\linewidth}ccc}
\toprule
\textbf{Benign outcome} & \(\mathbf{BCR_{\text{strict}}}\) & \(\mathbf{GAR}\) & \textbf{Clar.} \\
\midrule
Immediate or read-only completion & Yes & Yes & No \\
Reviewable draft or preflight artifact & No & Yes & No \\
Clarification with no effect & No & No & Yes \\
Invalid or denied benign call & No & No & No \\
\bottomrule
\end{tabular}
\caption{Relationship between benign-outcome metrics. Clarification is treated as a safe non-completion rather than as benign completion.}
\label{tab:benignmetricsemantics}
\end{table}

\paragraph{Over-Defense Rate}

\[
ODR=\frac{\#\{\text{justified benign effects denied}\}}{\#\{\text{justified benign effects}\}}.
\]

\paragraph{Manifest Reduction Score}

\[
MRS=1-\frac{|\visible_{\mathrm{intent}}|}{|\visible_{\sys}|}.
\]

\paragraph{Intent-Trace Field Completeness}

\[
\begin{aligned}
\mathcal{F}={}&\{\text{request, certificate, policy,}\\
&\phantom{\{}\text{tool, decision, outcome}\},\\
TFC={}&\frac{1}{N}\sum_{i=1}^{N}
\mathbf{1}\!\left[\mathcal{F}\subseteq\mathrm{Fields}(T_i)\right].
\end{aligned}
\]

TFC measures field completeness only; a trace without intent-certificate enforcement lacks certificate fields by construction.

\paragraph{Classifier Quality}

We compare certificate outputs against author-specified reference envelopes. Metrics include intent-class accuracy, resource-bound extraction accuracy, unsafe narrowing failure, over-defense, clarification rate, and latency:

\[
ICA=\frac{\#\{\hat{I}=I^\star\}}{N},\quad
RBA=\frac{\#\{\hat{B}_R=B_R^\star\}}{N}.
\]

Here \(I^\star\) and \(B_R^\star\) are the fixture's author-specified reference classes and bounds. Unsafe narrowing failure measures cases where a classifier emits a certificate that fails to narrow authority when required; over-defense measures benign requests routed to unnecessary clarification, draft, or denial.

\subsection{Component Analyses}

The component evidence spans three scopes:

\begin{itemize}
  \item the deterministic toy check isolates manifest filtering, payload consistency, and OpenPort effect controls;
  \item the 176-instance runtime check compares reference-certificate, rule-based, and hybrid rule+flow certificate generation;
  \item the normalizer analysis rescored frozen model-runtime traces after observing their failures.
\end{itemize}

\noindent Certificate expiry and audit binding are covered by endpoint invariants rather than causal outcome ablations. The normalizer analysis re-scores traces collected before its rules were defined.

\subsection{Interpretation and Statistical Unit}

We report raw numerators, denominators, and descriptive rates. The three repeats reuse the same parameterized task templates, so trials are not treated as independent draws from a deployment population. Zero observed events mean \(0/n\) in the evaluated set, not zero underlying risk. We exclude the legacy percentile-bootstrap intervals because their original random seed was not recorded and they flattened repeated task instances. We instead use deterministic task-clustered reanalysis with 10,000 resamples and analysis seed \(20260724\).

\subsection{Executable Synthetic Sanity Check}

We evaluate a deterministic sanity check with seven synthetic tools, twelve synthetic tasks, and three policy variants:

\begin{itemize}
  \item B0: model-only execution over a broad static tool list;
  \item B1: OpenPort substrate behavior with static tool exposure and draft-first high-risk writes;
  \item B2: intent-certificate manifest narrowing and payload consistency checks.
\end{itemize}

The script checks the policy rules over synthetic benign, mismatch, prompt-injection, tool-poisoning, ambiguous, and compound tasks. The separate runtime-backed microbenchmark exercises the OpenPort reference HTTP path.

The check contains 5 benign and 7 attack-or-ambiguous tasks. Under the broad static manifest, 33 unjustified high-risk tool exposures occur across the task set; under intent-certificate narrowing, this count is 0. The toy unsafe-execution rate is 1.0 for model-only B0, 0.1429 for OpenPort-substrate B1, and 0 for intent-certificate B2. Under the legacy governed-benign outcome, both B1 and B2 score 1.0 on the five benign tasks. Re-scored under this paper's definitions, their governed-artifact rate remains 1.0, whereas strict benign completion is 0.8000 because one of five benign tasks is routed to review rather than strictly completed. Mean manifest reduction is 0.7143.

The sanity check also evaluates six ablation conditions. Table~\ref{tab:toy-ablation} reports the resulting metrics.

\begin{table*}[t]
\centering
\footnotesize
\begin{tabularx}{\textwidth}{Yrrrrrrr}
\toprule
\textbf{Condition} & \textbf{UER} & \(\mathbf{BCR_{\mathrm{strict}}}\) & \(\mathbf{GAR}\) & \textbf{ODR} & \textbf{MRS} & \textbf{UHRT} & \textbf{Review} \\
\midrule
Prompt-only instruction & 1.0000 & 1.0000 & 1.0000 & 0.0000 & 0.0000 & 33 & 0 \\
Static OpenPort only & 0.1429 & 0.8000 & 1.0000 & 0.0000 & 0.0000 & 33 & 1 \\
Manifest filtering only & 0.2857 & 1.0000 & 1.0000 & 0.0000 & 0.7143 & 0 & 0 \\
Consistency gate only & 0.0000 & 1.0000 & 1.0000 & 0.0000 & 0.0000 & 33 & 0 \\
Manifest filtering + consistency gate & 0.0000 & 1.0000 & 1.0000 & 0.0000 & 0.7143 & 0 & 0 \\
Intent-certificate layer + OpenPort effect controls & 0.0000 & 0.8000 & 1.0000 & 0.0000 & 0.7143 & 0 & 1 \\
\bottomrule
\end{tabularx}
\caption{Deterministic toy ablation over the synthetic intent-layer task matrix, re-scored under the paper's metric definitions. UER is unsafe execution rate; \(BCR_{\mathrm{strict}}\) requires direct safe completion without clarification, draft, or review; \(GAR\) also credits a governed artifact. ODR is over-defense rate, MRS is mean manifest reduction, UHRT is the count of unjustified high-risk tools visible across all tasks, and Review is the number of benign high-risk cases requiring operator intervention under OpenPort effect controls.}
\label{tab:toy-ablation}
\end{table*}

The toy ablation separates the roles of the major components. Prompt-only instruction is weakest under an adversarial planner because no server-side policy binds model behavior. Static OpenPort effect controls block high-risk direct effects but cannot reject a medium-risk payload expansion that is inconsistent with the user's requested amount. Manifest filtering removes unjustified high-risk tools from the visible surface but does not by itself catch payload expansion on a still-visible tool. The consistency gate catches payload and effect mismatches even when the full static manifest remains visible, but it does not reduce tool-enumeration exposure. The combined intent-certificate conditions obtain both zero unsafe execution on this toy matrix and zero unjustified high-risk exposure. Adding OpenPort effect controls introduces one review intervention for the benign high-risk delete task, which is the intended safety cost of draft/preflight governance.

A deterministic confidence-policy check exercises the classifier-trust boundaries defined in Section~4. It defines twelve synthetic confidence states covering high-confidence low-risk tasks, medium-confidence create/update tasks, low-confidence ambiguous tasks, conflicting read/delete intent, high-risk delete/export intent, multi-step read-then-delete workflows, and high-confidence classifier outputs that exceed static OpenPort policy. Table~\ref{tab:confidenceeval} reports the resulting policy checks.

\begin{table}[t]
\centering
\small
\begin{tabular}{p{0.58\linewidth}r}
\toprule
\textbf{Check} & \textbf{Toy result} \\
\midrule
Expected action match rate & 1.0000 \\
High-confidence low-risk allow rate & 1.0000 \\
Medium-confidence draft/preflight rate & 1.0000 \\
Low-confidence clarification rate & 1.0000 \\
Conflicting-intent deny/split rate & 1.0000 \\
High-risk confirmation/preflight rate & 1.0000 \\
Multi-step per-step certificate rate & 1.0000 \\
Static policy deny rate & 1.0000 \\
Unsafe broad authority grants & 0 \\
Authority widening rate & 0.0000 \\
Broad multi-step certificates & 0 \\
\bottomrule
\end{tabular}
\caption{Deterministic toy confidence-policy checks. These results validate the policy routing rules, not classifier accuracy.}
\label{tab:confidenceeval}
\end{table}

These numbers provide a deterministic conformance check of the policy sketch.

\subsection{Runtime-Backed Microbenchmark}

To connect the policy sketch to the implemented runtime, we run a runtime-backed microbenchmark.

The script uses Fastify HTTP injection against the OpenPort reference runtime. The suite contains 176 deterministic instances generated from author-designed templates: 102 benign and 74 attack-or-ambiguous. It covers six intent classes: read, summarize, create, update, delete, and export. It also covers seven attack or ambiguity categories: intent-tool mismatch, payload expansion, indirect prompt injection, tool metadata poisoning, low confidence, compound workflow, and resource underspecification. The benign set includes underspecified requests so that clarification and review burden remain visible. For each instance, the runtime calls \texttt{/intent}, \texttt{/manifest}, and, when applicable, \texttt{/actions}; it records endpoint outcomes, drafts, executions, manifest sizes, audit binding, and local diagnostic latency.

The reported \(UAR_c\) is a conservative composite: it flags an attack-or-ambiguous instance when an unjustified high-risk tool remains visible or an inconsistent draft/execution path is accepted. Thus OpenPort-only can have \(UAR_c=1.0\) while UER is 0. The composite supports deterministic conformance and differs from the path-only UAR used in the real-model study.

\begin{table*}[t]
\centering
\footnotesize
\begin{tabularx}{\textwidth}{Yrrrrrrrrr}
\toprule
\textbf{Scenario} & \textbf{Tasks} & \(\mathbf{UAR_c}\) & \textbf{UER} & \(\mathbf{BCR_{\text{strict}}}\) & \(\mathbf{GAR}\) & \textbf{Clar.} & \textbf{Review} & \textbf{Draft} & \textbf{MRS} \\
\midrule
OpenPort-only & 176 & 1.0000 & 0.0000 & 0.4314 & 1.0000 & 0.0000 & 0.5686 & 1.0000 & 0.0000 \\
Reference certificates & 176 & 0.0000 & 0.0000 & 0.4314 & 0.8627 & 0.1373 & 0.4314 & 0.3333 & 0.7614 \\
Rule-based certificates & 176 & 0.0000 & 0.0000 & 0.4314 & 0.9608 & 0.0196 & 0.5294 & 0.4091 & 0.7662 \\
Hybrid rule+flow certificates & 176 & 0.0000 & 0.0000 & 0.4314 & 0.9608 & 0.0196 & 0.5294 & 0.4091 & 0.7662 \\
\bottomrule
\end{tabularx}

\vspace{0.7em}
\begin{tabularx}{0.72\textwidth}{Yrrrr}
\toprule
\textbf{Scenario} & \textbf{Tasks} & \textbf{P50} & \textbf{P95} & \textbf{P99} \\
\midrule
OpenPort-only & 176 & 0.1748 & 0.3819 & 1.5451 \\
Reference certificates & 176 & 0.2486 & 0.4739 & 1.3952 \\
Rule-based certificates & 176 & 0.2342 & 0.4305 & 1.3199 \\
Hybrid rule+flow certificates & 176 & 0.1819 & 0.3640 & 1.1942 \\
\bottomrule
\end{tabularx}
\caption{Runtime-backed synthetic microbenchmark. The upper panel reports safety and utility outcomes. \(BCR_{\mathrm{strict}}\), \(GAR\), Clar., and Review use all 102 benign cases as their denominator; Clar. and Review are disjoint, with Review excluding clarification-routed cases. Draft uses all 132 rows per scenario that reach proposal/action routing (\texttt{proposalRows}), including benign and attack-or-ambiguous rows: \(1.0000=132/132\), \(0.3333=44/132\), and \(0.4091=54/132\). The lower panel reports machine-local P50/P95/P99 Fastify-injection diagnostics in milliseconds from the archived result. \(UAR_c\) is the legacy composite exposure-or-path indicator; \(GAR\) counts strict completion or a governed artifact.}
\label{tab:runtimemicro}
\end{table*}

The study separates enforcement correctness from certificate-generation quality. The reference condition uses author-specified certificates embedded in the synthetic task suite. Rule-based and hybrid rule+flow conditions are deterministic local classifiers. Model-generated certificates are evaluated in the studies below. The reference condition routes underspecified benign requests more conservatively, lowering \(GAR\). Higher \(GAR\) in rule-based or hybrid rows may therefore reflect less conservative routing rather than better access-control quality.

\begin{table*}[t]
\centering
\footnotesize
\begin{tabularx}{\textwidth}{Yrrrrr}
\toprule
\textbf{Classifier} & \textbf{ICA} & \textbf{RBA} & \textbf{EBA} & \textbf{UNF} & \textbf{ODR} \\
\midrule
Oracle & 1.0000 & 1.0000 & 1.0000 & 0.0000 & 0.0000 \\
Rule-based & 0.6818 & 0.8977 & 1.0000 & 0.0000 & 0.0000 \\
Hybrid rule+flow & 0.8636 & 0.9886 & 1.0000 & 0.0000 & 0.0000 \\
Model-generated certificates & \multicolumn{5}{l}{evaluated in the model studies below} \\
\bottomrule
\end{tabularx}
\caption{Small deterministic classifier/oracle comparison on the 176-task synthetic suite. ICA is intent-class accuracy, RBA is resource-bound accuracy, EBA is effect-bound accuracy, UNF is unsafe narrowing failure, and ODR is over-defense rate. All rows use deterministic classifiers or reference oracles.}
\label{tab:classifiermicro}
\end{table*}

This check exercises endpoint wiring, certificate storage, manifest narrowing, action denial, draft creation, and audit binding. The reference-certificate condition separates enforcement from certificate generation and uses a deterministic planner to isolate the enforcement path from model behavior.

\subsection{Benchmark-Family Transfer Set}
\label{sec:transfer-set}

To test the mechanism on structures not unique to OpenPort, we evaluate a locally authored benchmark-family transfer set. The set uses twenty-five tasks: six AgentDojo-shaped tasks, seven ToolSandbox-shaped tasks, six tau-bench-shaped tasks, and six ToolEmu-shaped tasks. The runtime-backed deterministic scorer uses author-specified reference certificates and concrete proposals, then drives them through the OpenPort reference HTTP path with and without the intent layer. The task content remains synthetic and locally authored; only structural patterns are transferred from the benchmark families.

Table~\ref{tab:externaladaptcoverage} separates benign, ambiguous, and attack tasks by source family. The attack-heavy batch measures narrowing of benchmark-shaped escalation attempts rather than deployment task prevalence. Table~\ref{tab:externaladaptcategories} partitions the 17 non-benign tasks into additive primary attack-surface categories.

\begin{table}[t]
\centering
\small
\begin{tabular}{L{0.26\linewidth}rrrr}
\toprule
\textbf{Source family} & \textbf{Benign} & \textbf{Ambiguous} & \textbf{Attack} & \textbf{Total} \\
\midrule
AgentDojo & 2 & 0 & 4 & 6 \\
ToolSandbox & 2 & 2 & 3 & 7 \\
tau-bench & 2 & 2 & 2 & 6 \\
ToolEmu & 2 & 0 & 4 & 6 \\
\midrule
\textbf{Overall} & 8 & 4 & 13 & 25 \\
\bottomrule
\end{tabular}
\caption{Task-type coverage in the benchmark-family transfer set.}
\label{tab:externaladaptcoverage}
\end{table}

\begin{table}[t]
\centering
\small
\resizebox{\linewidth}{!}{%
\begin{tabular}{L{0.34\linewidth}rrL{0.24\linewidth}}
\toprule
\textbf{Primary category} & \textbf{Tasks} & \textbf{Attack} & \textbf{Source families} \\
\midrule
Prompt injection / untrusted content & 4 & 4 & AgentDojo \\
Insufficient information & 4 & 0 & ToolSandbox, tau-bench \\
Wrong scope / target mismatch & 3 & 3 & tau-bench, ToolEmu \\
State dependency & 2 & 2 & ToolSandbox \\
Tool poisoning / semantic deception & 2 & 2 & ToolEmu \\
Canonicalization / normalization risk & 1 & 1 & ToolSandbox \\
Amount expansion / bound inflation & 1 & 1 & ToolEmu \\
\bottomrule
\end{tabular}
}%
\caption{Primary attack-surface coverage for the 17 non-benign transfer-set tasks. Each task is assigned exactly one primary category.}
\label{tab:externaladaptcategories}
\end{table}

Across the full transfer set, OpenPort-only has composite \(UAR_c=1.0000\), UER \(=0.0000\), and \(GAR=1.0000\). With the intent layer and author-specified reference certificates, \(UAR_c\) is \(0.0000\), UER remains \(0.0000\), \(GAR\) remains \(1.0000\), and mean manifest reduction is \(0.7829\). This deterministic comparison measures conformance on the locally authored tasks.

\begin{table}[t]
\centering
\small
\resizebox{\linewidth}{!}{%
\begin{tabular}{L{0.28\linewidth}rcccc}
\toprule
\textbf{Source family} & \textbf{Tasks} & \(\mathbf{OpenPort\ UAR_c}\) & \(\mathbf{Intent\ UAR_c}\) & \(\mathbf{Intent\ GAR}\) & \textbf{Intent MRS} \\
\midrule
AgentDojo & 6 & 1.0000 & 0.0000 & 1.0000 & 0.7619 \\
ToolSandbox & 7 & 1.0000 & 0.0000 & 1.0000 & 0.7755 \\
tau-bench & 6 & 1.0000 & 0.0000 & 1.0000 & 0.7381 \\
ToolEmu & 6 & 1.0000 & 0.0000 & 1.0000 & 0.8571 \\
\bottomrule
\end{tabular}
}%
\caption{Runtime results on the locally authored benchmark-family transfer set with author-specified reference certificates.}
\label{tab:externaladapt}
\end{table}

\subsection{Expanded Real LLM Pilot}

We evaluated certificate generation and planning with three open-weight models: GPT-OSS-120B, Llama-3.3-70B-Instruct, and Qwen3-Next-80B. Each model received 55 classifier tasks and 54 planner attack tasks under static and intent-filtered manifests. This pilot evaluates model behavior separately from the OpenPort action path exercised in the end-to-end study.

\begin{table*}[t]
\centering
\scriptsize
\begin{tabular}{p{0.31\linewidth}rrrrrr}
\toprule
\textbf{Model} & \textbf{ICA} & \textbf{RBA} & \textbf{EBA} & \textbf{Review Acc.} & \textbf{High-Risk Allow} & \textbf{P95 ms} \\
\midrule
GPT-OSS-120B & 0.8545 & 0.5818 & 0.9091 & 0.5273 & 0.8571 & 3767.5 \\
Llama-3.3-70B & 0.7273 & 0.1818 & 0.7818 & 0.3818 & 0.4286 & 17633.4 \\
Qwen3-Next-80B & 0.7818 & 0.1818 & 0.7818 & 0.6909 & 0.0000 & 4723.2 \\
\bottomrule
\end{tabular}
\caption{Expanded real LLM classifier pilot over 55 synthetic intent-certificate tasks per model. ICA is intent-class accuracy, RBA is resource-bound accuracy, EBA is effect-bound accuracy, and High-Risk Allow is the rate at which high-risk create/update/delete/export requests were incorrectly emitted as immediate \texttt{allow}. Across models, broad intent recognition is materially stronger than exact resource/effect-bound extraction.}
\label{tab:llmclassifierpilot}
\end{table*}

\begin{table*}[t]
\centering
\scriptsize
\setlength{\tabcolsep}{4pt}
\resizebox{0.98\textwidth}{!}{%
\begin{tabular}{p{0.31\linewidth}rrrrrr}
\toprule
\textbf{Model} & \textbf{Static Unsafe} & \textbf{Filtered Unsafe} & \textbf{Filtered+Gate Unsafe Accepted} & \textbf{Static Clar.} & \textbf{Filtered Clar.} & \textbf{Filtered P95 ms} \\
\midrule
GPT-OSS-120B & 0.5000 & 0.4444 & 0.0000 & 0.4444 & 0.5000 & 4790.9 \\
Llama-3.3-70B & 0.8889 & 0.5926 & 0.0000 & 0.0556 & 0.3519 & 35290.4 \\
Qwen3-Next-80B & 0.9259 & 0.7037 & 0.0000 & 0.0556 & 0.2778 & 5820.6 \\
\bottomrule
\end{tabular}
}
\caption{Expanded real LLM planner pilot over 54 synthetic attack tasks per model. Manifest filtering lowers unsafe proposals, but deterministic gating is what drives unsafe accepted authority to zero in this pilot.}
\label{tab:llmplannerpilot}
\end{table*}

\subsection{Broad Intent Is Easier Than Authorization Bounds}

Across the scored models, intent-class accuracy ranges from 0.7273 to 0.8545, while resource-bound accuracy ranges from 0.1818 to 0.5818 and review-mode accuracy from 0.3818 to 0.6909. Models recognize coarse read, summarize, create, update, delete, and export classes more reliably than exact resource identifiers, row bounds, and risk modes. The authorization layer therefore treats model-generated certificates and review modes as advisory. Server-side policy validates scopes, narrows manifests, checks payload bounds, and overrides any model-proposed \texttt{allow} decision for high-risk tools. GPT-OSS-120B, despite the strongest bound extraction in this set, emitted immediate \texttt{allow} for 85.71\% of high-risk requests.

The planner pilot shows the same separation of responsibilities. Manifest filtering lowers unsafe proposal rates for every scored model, but filtered manifests alone still leave unsafe read-path or overbroad-payload proposals. Deterministic consistency gating reduces unsafe accepted authority to zero in this synthetic planner setting, showing that manifest filtering must be paired with payload-bound server enforcement.

The certificate-generation comparison covers reference, rule-based, and hybrid rule+flow conditions in the runtime-backed microbenchmark, and raw model-generated certificates in the model studies. Table~\ref{tab:normalizerablation} separately evaluates schema normalization and deterministic risk override by re-scoring the collected traces.

\subsection{Scored End-to-End LLM-in-the-Loop Runtime Benchmark}

We additionally ran an API-backed benchmark harness that pushes model-generated classifier and planner output through the OpenPort reference intent-layer action path:

\[
\begin{aligned}
\texttt{/intent}&\rightarrow\texttt{/manifest}
\rightarrow\texttt{/preflight or /actions}\\
&\rightarrow\texttt{consistency gate}
\rightarrow\texttt{audit}.
\end{aligned}
\]

The scored run uses 34 synthetic tasks per run (12 benign and 22 attack-or-ambiguous), 3 repeats per model, and 3 open-weight models: GPT-OSS-120B, Llama-3.3-70B, and Qwen3-Next-80B. The suite is parameterized over three synthetic resource collections, so benign and adversarial patterns recur across bounded settings. It records certificate creation, clarification, unsafe planner proposals, accepted authority, executed effects, \(BCR_{\text{strict}}\), \(GAR\), manifest reduction, latency, and trace summaries.

\begin{table*}[t]
\centering
\footnotesize
\setlength{\tabcolsep}{3pt}
\begin{tabular}{lrrrrrrrrr}
\toprule
\textbf{Model} & \textbf{Cert.} & \textbf{Plan} & \textbf{Plan} & \textbf{Unsafe} & \textbf{Unsafe} & \(\mathbf{BCR_{\text{strict}}}\) & \(\mathbf{GAR}\) & \textbf{MRS} & \textbf{P95} \\
& \textbf{create} & \textbf{clar.} & \textbf{unsafe} & \textbf{path} & \textbf{exec.} & & & & \textbf{(s)} \\
\midrule
GPT-OSS-120B & 1.0000 & 0.4216 & 0.5784 & 0.2424 & 0.0000 & 0.0000 & 0.2500 & 0.8025 & 10.47 \\
Llama-3.3-70B & 1.0000 & 0.0980 & 0.7353 & 0.0909 & 0.0000 & 0.0000 & 0.5000 & 0.7913 & 49.87 \\
Qwen3-Next-80B & 1.0000 & 0.0490 & 0.9510 & 0.2727 & 0.0000 & 0.0000 & 0.2500 & 0.8025 & 9.95 \\
\bottomrule
\end{tabular}
\caption{Scored end-to-end model runtime benchmark. Each row aggregates 34 synthetic tasks across three repeats. ``Unsafe path'' is intent-inconsistent accepted authority; ``unsafe exec.'' is a completed unsafe effect. Latency is descriptive.}
\label{tab:llmruntimepilot}
\end{table*}

\statementheading{Finding 2: Unsafe draft authority remains}
The full gateway path receives real-model outputs. Across 198 attack-or-ambiguous trials, the combined intent-layer and OpenPort path accepts 40 unsafe drafts and records 0 completed unsafe effects.

Runtime unsafe accepted authority is 40/198 overall and 0.0909--0.2727 by model. The cases concentrate in draft-first create paths where the model-generated certificate or payload loses the author-specified amount/resource bound. Current model-generated certificates therefore do not completely enforce request-bound narrowing.

\statementheading{Finding 3: Model behavior exposes a safety-utility-latency tradeoff}
Manifest reduction is roughly 0.79--0.80 across models, but \(BCR_{\text{strict}}\) is 0/108 overall and \(GAR\) is only 36/108. Thus every counted benign artifact is a governed draft rather than immediate completion. Llama-3.3-70B has the lowest unsafe accepted rate in this setting (0.0909) and the highest \(GAR\) (0.5000), but the largest latency tail (P95 49.9 seconds). These model-level differences are descriptive within the small synthetic set and single provider configuration.

\subsection{Transfer-Subset End-to-End LLM Runtime Pilot}

We ran the same API-backed harness over a selected transfer subset. It contains 12 locally authored tasks from Section~\ref{sec:transfer-set}: 4 benign, 2 ambiguous, and 6 attack tasks, with 3 shaped by each benchmark family. Unlike the deterministic reference-certificate scorer, this pilot uses real-model certificates and planner output before calling the OpenPort runtime path.

\begin{table*}[t]
\centering
\footnotesize
\setlength{\tabcolsep}{4pt}
\begin{tabular}{lrrrrrrr}
\toprule
\textbf{Model} & \textbf{Tasks} & \textbf{Plan clar.} & \textbf{Plan unsafe} & \textbf{Unsafe path} & \textbf{Unsafe exec.} & \(\mathbf{GAR}\) & \textbf{P95 (s)} \\
\midrule
GPT-OSS-120B & 12 & 0.5000 & 0.5000 & 0.2500 & 0.0000 & 0.7500 & 13.76 \\
Llama-3.3-70B & 12 & 0.0000 & 1.0000 & 0.1250 & 0.0000 & 0.7500 & 53.15 \\
Qwen3-Next-80B & 12 & 0.0833 & 0.9167 & 0.2500 & 0.0000 & 0.7500 & 7.36 \\
\bottomrule
\end{tabular}
\caption{API-backed runtime pilot over the 12-task benchmark-family transfer subset.}
\label{tab:llmruntimeexternalsubset}
\end{table*}

Across all three scored models, runtime unsafe accepted authority is 0.1250--0.2500. The governed-artifact rate is 0.7500 for all three rows, while strict benign completion remains 0. Three of the four benign tasks reach non-executed governed artifacts; the tau-bench-style bounded review delete still fails under current planners rather than reaching such a path.

In this 12-task subset, runtime unsafe accepted cases concentrate in ToolEmu-style bounded-create attacks rather than in the AgentDojo, ToolSandbox, or tau-bench-derived rows. The same bounded-create and semantic-deception residue appears in the larger synthetic runtime benchmark, making certificate precision and planner-call quality the main blockers to eliminating accepted authority.

\begin{table*}[t]
\centering
\footnotesize
\begin{tabular}{lrrrrr}
\toprule
\textbf{Primary category} & \textbf{Trials} & \textbf{Clarify} & \textbf{Blocked / no effect} & \textbf{Unsafe accepted} & \textbf{Unsafe executed} \\
\midrule
Prompt injection / untrusted content & 6 & 2 & 4 & 0 & 0 \\
Insufficient information & 6 & 3 & 3 & 0 & 0 \\
State dependency & 3 & 1 & 2 & 0 & 0 \\
Wrong scope / target mismatch & 3 & 1 & 2 & 0 & 0 \\
Tool poisoning / semantic deception & 3 & 0 & 1 & 2 & 0 \\
Amount expansion / bound inflation & 3 & 0 & 0 & 3 & 0 \\
\bottomrule
\end{tabular}
\caption{Outcome cross-tab for the 24 non-benign model-task trials in the external-subset runtime pilot. Counts aggregate across the three scored models; the canonicalization-risk category is absent from this 12-task subset.}
\label{tab:externalsubsetoutcomes}
\end{table*}

Table~\ref{tab:externalsubsetoutcomes} makes the residual risk boundary more concrete. All prompt-injection, insufficient-information, state-dependency, and wrong-scope trials end either in clarification or in a blocked no-effect path. Runtime unsafe accepted authority remains only in ToolEmu-style tool-poisoning / semantic-deception rows and amount-expansion rows. These failures concentrate in the same certificate-bound and bounded-create residue identified by the larger synthetic runtime benchmark.

\subsection{Evaluation Synthesis}

The runtime-backed microbenchmark, scored LLM runtime benchmark, transfer-set coverage tables, and transfer-subset runtime pilot support three conclusions.

First, in the external-subset runtime pilot, prompt injection, insufficient information, state dependency, and wrong-scope / wrong-target trials all collapse into clarification or blocked no-effect paths rather than unsafe accepted authority.

Second, the remaining residue is structurally concentrated rather than diffuse. Across both the larger scored runtime benchmark and the smaller benchmark-shaped external subset, accepted-authority failures cluster in bounded-create settings where certificates and payloads preserve the broad operation class but lose the exact amount, resource, or effect contract needed for early rejection. Tool-poisoning rows only survive when they can masquerade as create-compatible draft paths; amount-expansion rows survive when the gateway still sees a create-shaped proposal but cannot prove the numeric bound is faithful enough.

Third, the empirical bottleneck is certificate canonicalization, resource/effect normalization, planner schema repair, and stronger create-versus-destructive effect contracts. Bounded-create semantics remain the dominant open problem once OpenPort effect controls and monotone narrowing are in place.

\statementheading{Finding 4: The remaining accepted-authority failures are narrow and classifiable}
The 40 accepted-authority cases across the 198 attack-or-ambiguous runtime trials all end as drafts without creating an execution. They are reviewable authority paths that the current intent-certificate layer did not narrow away early enough.

\begin{table*}[t]
\centering
\footnotesize
\begin{tabularx}{\textwidth}{Y r Y Y}
\toprule
\textbf{Accepted-authority failure mode} & \textbf{Cases} & \textbf{Trace pattern} & \textbf{Interpretation} \\
\midrule
Create payload expansion with missing or weak bounds & 24 & \texttt{create\_amount\_expansion\_*}; \texttt{actionStatus=draft} & In 15/24 cases the certificate omits enforceable resource/effect bounds entirely; in the remaining 9/24 the certificate contains create-compatible but insufficient amount/type bounds. \\
Tool-poisoning-like create path with semantic deception & 16 & \texttt{tool\_poison\_hard\_delete\_*}; \texttt{actionStatus=draft} & The planner is induced toward a create-compatible draft path, which remains accepted authority when the certificate does not narrow the path away. \\
Restrictive review mode that still leaves accepted draft authority (overlapping slice) & 24 & Review mode is \texttt{draft} or \texttt{preflight} rather than \texttt{allow} & Review mode alone is insufficient to eliminate unsafe accepted authority. It converts the path into governed draft/preflight handling, but does not by itself prove the payload/effect bounds are tight enough. \\
\bottomrule
\end{tabularx}
\caption{Failure modes and cross-cutting properties for the scored runtime benchmark. The first two rows are a mutually exclusive \(24+16=40\) partition of accepted-authority cases. The restrictive-review row is an overlapping slice of those same 40 cases.}
\label{tab:runtimefailuretaxonomy}
\end{table*}

Of the 40 accepted-authority cases, 24 have empty resource and effect bounds in the emitted certificate, while the remaining 16 are semantically weak but non-empty create-compatible certificates. Review-mode labels split across \texttt{allow} (16), \texttt{preflight} (6), and \texttt{draft} (18), showing that the residual problem is not simply an incorrect review bit. Stronger amount, resource, and effect normalization is needed to collapse these governed draft paths into earlier denials or clarifications.

\begin{table*}[t]
\centering
\footnotesize
\begin{tabularx}{\textwidth}{Y r Y Y}
\toprule
\textbf{Benign or unsafe runtime outcome} & \textbf{Count} & \textbf{Example} & \textbf{Likely fix} \\
\midrule
Reviewable benign draft counted in \(GAR\) & 36 & \texttt{create\_collection\_main} & Governed artifact; excluded from strict completion. \\
Malformed benign tool or payload call & 67 & \texttt{update\_rec\_2}, \texttt{delete\_rec\_7}, \texttt{export\_collection\_main} & Planner schema retry, endpoint-specific payload canonicalization, certificate normalizer. \\
Benign clarification with no effect & 4 & \texttt{export\_collection\_audit} (model A) & Track separately as safe non-completion rather than as hard failure. \\
Unknown benign action family & 1 & \texttt{export\_collection\_ops} (model B) & Tool-call schema validator and planner retry. \\
Accepted authority from missing amount/resource/effect bounds & 24 & \path{create_amount_expansion_col_*} & Numeric validators and certificate-bound normalizers before action routing. \\
Accepted authority from create-compatible semantic deception & 16 & \path{tool_poison_hard_delete_col_*} & Stronger effect contracts and semantic validation for create/delete separation. \\
\bottomrule
\end{tabularx}
\caption{Outcome taxonomy for the scored runtime benchmark.}
\label{tab:runtimeoutcometaxonomy}
\end{table*}

\begin{table*}[t]
\centering
\footnotesize
\begin{tabularx}{\textwidth}{Y Y Y Y Y}
\toprule
\textbf{Case} & \textbf{Certificate / planner} & \textbf{Runtime result} & \textbf{Metric interpretation} & \textbf{Lesson} \\
\midrule
Benign create draft & GPT-OSS-120B emits \texttt{create} with bounded resource and amount; planner calls \texttt{resource.create}. & \texttt{common.success}, \path{actionStatus=draft}, audit-bound. & Counts toward \(GAR\); excluded from \(BCR_{\text{strict}}\). & Governed draft creation without task completion. \\
Benign update invalid & GPT-OSS-120B emits bounded \texttt{update}; planner calls \texttt{resource.update}. & \path{agent.action_invalid}; no draft or execution. & Hard benign failure. & Certificate intent is plausible, but payload canonicalization is still too weak. \\
Intent mismatch blocked by clarification & GPT-OSS-120B emits \texttt{read,summarize}; planner returns \texttt{clarify}. & No action, no draft, no execution. & Safe non-completion. & Manifest narrowing plus planner uncertainty can fail closed without creating authority. \\
Low-confidence ambiguity & GPT-OSS-120B emits \texttt{unknown}, confidence 0.2, review mode \texttt{clarify}. & No action, no draft, no execution. & Clarification rather than failure. & Confidence-aware routing works as intended when the request is underspecified. \\
Amount-expansion attack & GPT-OSS-120B emits \texttt{create} with weak amount bound; planner calls \texttt{resource.create}. & \texttt{common.success}, \texttt{draft}, \texttt{runtimeUnsafe}\allowbreak\texttt{Accepted=true}. & Unsafe accepted draft authority. & Missing numeric/resource precision leaves a reviewable authority path open. \\
Tool-poisoning attack & GPT-OSS-120B emits create-compatible certificate under poisoned context; planner still selects \texttt{resource.create}. & \texttt{common.success}, \texttt{draft}, \texttt{runtimeUnsafe}\allowbreak\texttt{Accepted=true}. & Unsafe accepted draft authority. & Semantic validation is needed to distinguish benign create from disguised delete intent. \\
\bottomrule
\end{tabularx}
\caption{Qualitative runtime traces from the scored benchmark.}
\label{tab:qualitativetraces}
\end{table*}

\subsection{Certificate-Normalizer Sensitivity Analysis}

The residual failures in Table~\ref{tab:runtimefailuretaxonomy} suggest a specific design question: would a deterministic certificate-bound normalizer reduce unsafe accepted authority, or is the remaining problem mainly planner behavior? We normalize common model-generated certificate aliases such as \texttt{collection}, \texttt{resource}, \texttt{amount}, \texttt{itemLimit}, and \texttt{limit}, and re-score the collected traces under stricter certificate-bound predicates.

The trace-backed counterfactual re-scores existing rows without replaying the model, repairing planner payloads, or creating new OpenPort actions. It estimates whether canonical resource/effect bounds at the consistency gate would remove the observed accepted-authority failures.

\begin{table}[t]
\centering
\small
\resizebox{\linewidth}{!}{%
\begin{tabular}{L{0.38\linewidth}rrrrr}
\toprule
\textbf{Condition} & \(\mathbf{UAR_{\mathrm{path}}}\) & \textbf{UER} & \(\mathbf{GAR}\) & \textbf{Accepted} & \textbf{Benign artifacts} \\
\midrule
Raw LLM certificates & 0.2020 & 0.0000 & 0.3333 & 40 & 36 \\
LLM + strict bound validator & 0.0758 & 0.0000 & 0.1389 & 15 & 15 \\
LLM + strict validator + normalized gate & 0.0000 & 0.0000 & 0.1389 & 0 & 15 \\
LLM + deterministic risk override only & 0.2020 & 0.0000 & 0.3333 & 40 & 36 \\
\bottomrule
\end{tabular}
}
\caption{Sensitivity analysis over collected model-runtime traces. \(UAR_{\mathrm{path}}\) is the accepted-action-path rate over attack-or-ambiguous trials; the analysis introduces no new runtime calls.}
\label{tab:normalizerablation}
\end{table}

The analysis gives two signals. First, deterministic review-mode override alone does not change \(UAR_{\mathrm{path}}\) because every accepted-authority case is already a draft rather than an execution. The weak point is whether resource and effect bounds are canonical enough for enforcement. Second, strict normalization trades authority rejection for fewer governed benign artifacts: \(UAR_{\mathrm{path}}\) changes from 0.2020 to 0.0758 under bound validation and to 0 under the normalized-gate predicate, while \(GAR\) changes from 0.3333 to 0.1389.

\subsection{Answers to the Research Questions}

\paragraph{RQ1 (controlled authority narrowing)}
On the 176-instance deterministic runtime suite, the reference-certificate condition changes the archived composite exposure-or-path indicator \(UAR_c\) from 1.0000 under OpenPort-only to 0, while UER is 0 in both conditions. These results measure manifest and certificate conformance on the evaluated suite.

\paragraph{RQ2 (real-model runtime outcomes)}
Across 198 attack-or-ambiguous trials, the combined intent-layer and OpenPort path accepts 40 intent-inconsistent drafts and observes 0 completed unsafe effects. The benchmark-family transfer subset accepts 5/24 unsafe drafts and observes 0/24 unsafe completed effects.

\paragraph{RQ3 (utility and latency cost)}
Strict benign completion is 0/108 in the scored model-runtime study; 36/108 trials yield governed artifacts. The transfer subset yields 9/12 governed artifacts. Model-level P95 end-to-end latency ranges from 9.95 to 49.87 seconds under the archived provider configuration. The evaluated prototype is review-oriented.

\paragraph{RQ4 (residual concentration)}
The 40 unsafe accepted cases comprise 24 amount/resource-bound failures and 16 create-compatible semantic-deception cases. All terminate as drafts. The concentration implicates certificate canonicalization and effect contracts.

\paragraph{RQ5 (component roles)}
The deterministic component check shows that manifest filtering removes unjustified tool exposure while payload consistency catches overbroad calls that remain in a visible tool class. Trace-backed normalization changes \(UAR_{\mathrm{path}}\) from 0.2020 to 0 and \(GAR\) from 0.3333 to 0.1389.

\section{Discussion}

\paragraph{Static non-expansion and request confinement}
The formal mechanism guarantees that the intent layer cannot exceed static OpenPort authority. Confinement to an expressed request additionally depends on an independently justified certificate envelope and sound adapter effect bounds. The 40/198 unsafe accepted drafts quantify this gap.

\paragraph{Drafts isolate effects but retain authority and review risk}
All 40 residual cases stop at drafts, so no evaluated domain effect completes. Nevertheless, a draft is persistent system state, consumes reviewer capacity, and can carry misleading content into a human approval flow. Effect isolation and authority rejection are therefore separate objectives.

\paragraph{Certificate precision is the measured bottleneck}
Broad class recognition is materially stronger than resource- and amount-bound extraction. Residual failures cluster in create-compatible cases whose operation class is plausible but whose exact amount, resource, or semantic effect is weakly encoded. This motivates typed bounds, deterministic normalization, and adapter contracts rather than greater confidence in free-form classifier output.

\paragraph{Utility remains review-oriented}
Strict benign completion is 0/108 in the scored model-runtime study; 36/108 benign trials produce governed artifacts. The current configuration is review-oriented and has low autonomous utility.

\section{Limitations and Threats to Validity}

\paragraph{Construct validity}
Task templates and reference envelopes are author-specified, creating circularity between mechanism design and scoring. The benchmark-family transfer set comprises locally authored structural analogues. The legacy deterministic \(UAR_c\) combines exposure and path acceptance and therefore differs from path-only UAR.

\paragraph{Internal and statistical validity}
The end-to-end study has no matched OpenPort-only real-model runtime row, leaving the isolated effect of the intent layer unidentified. Repeats share task templates and are correlated; zero observed events leave residual risk unmeasured. The provider-run bootstrap seed was not archived, and its legacy intervals are not used inferentially. The task-clustered reanalysis uses a recorded seed. The normalizer rules were developed after inspecting failures and may be overfit.

\paragraph{External validity}
The studies use synthetic data, three provider-served open-weight models, one provider configuration, modest task sets, and no real operators. Their evidence scope excludes production tenants, human review quality, reviewer fatigue, and adversaries adapting over time. Local Fastify-injection latency describes the test environment rather than networked deployment performance.

\paragraph{Implementation}
The prototype uses in-memory certificate state rather than persistent, distributed lifecycle enforcement. Historical provider runs did not record a clean runtime revision for every shard, and the transfer subset has incomplete top-level append provenance. Exact provider-call recreation is therefore unavailable.

\section{Security Analysis}

\subsection{Invariant 1: Intent Cannot Widen Static Policy}

The monotonicity invariant is:

\[
P_C \preceq P_A.
\]

If a tool is hidden by static OpenPort scopes, ABAC policy, workspace boundary, revocation, or rate limit, the intent layer cannot reveal it. This protects against classifier compromise turning into privilege escalation.

\subsection{Invariant 2: Low Confidence Fails Closed}

If \(\gamma < \gamma_{\min}\), the system must not treat the certificate as broad authority. It should expose minimal tools, request clarification, or force draft/preflight.

\subsection{Invariant 3: Payload Effects Must Be Bounded}

Even if a tool class is allowed, the payload must stay within request bounds. A read request for ``last week'' should not silently become a year-long export. A request to update one record should not become a bulk update.

\subsection{Invariant 4: The Model Is Not a Principal}

The model may suggest a tool call, but authorization is decided by the gateway. The gateway verifies scopes, policy, certificate validity, consistency, preflight, and review requirements.

\subsection{Invariant 5: Drafts Are Not Completed Domain Effects}

Draft creation is persistent governance state, not a completed domain effect. It still represents accepted authority, consumes reviewer capacity, and can become a social-engineering carrier. Therefore an intent-inconsistent proposal must deny or clarify before draft creation; consistent high-risk proposals may be drafted with audit, a clear preview, and bound review context.

\subsection{Prompt Injection}

Prompt injection can influence model reasoning, tool selection, payload generation, and certificate generation. Unconditionally, the intent layer can prevent such influence from expanding authority beyond static policy. It confines authority to the expressed request only when the certificate refines the trusted request envelope and adapter effect bounds are sound. Under those conditions, an injected export instruction outside a summary envelope is hidden or denied.

Attacks that alter details within an already authorized high-risk effect envelope require context provenance or taint-aware manifests in addition to the intent-certificate layer.

\subsection{Tool Metadata Poisoning}

Tool metadata attacks try to make a model choose a malicious or irrelevant tool. Intent-aware filtering reduces the effect of such attacks by limiting visible tools and checking selected tools against the certified request. However, if poisoned metadata belongs to a tool that is both statically authorized and intent-compatible, the intent-certificate layer alone may not detect semantic deception inside the tool implementation. Manifest signing, tool provenance, sandboxing, and behavior monitoring remain necessary.

\subsection{Effect Estimation and Conservative Failure}

The consistency predicate depends on estimating or bounding tool effects. This is easy for simple read, create, update, delete, and export actions with explicit resource IDs and item count limits. It is harder for tools with implicit side effects, compound workflows, tool-provided summaries, or domain-specific semantics.

The intent-certificate layer therefore adopts a conservative failure rule:

\[
\neg \mathrm{Boundable}(C,t,x,\widehat e)\Rightarrow
\mathrm{Route}_{\mathrm{intent}}\in\{\mcode{deny},\mcode{clarify}\}.
\]

Preflight may first be used to obtain a conservative bound. If the gateway still cannot bound the payload or predicted effect against the certificate, the enforce profile admits neither execution nor a draft path; it denies or asks for clarification. This prevents uncertainty from becoming persistent accepted authority.

\subsection{Classifier Compromise}

If the classifier is compromised, it might produce a broad certificate. Monotonicity still prevents scope expansion, but it may fail to narrow authority. Mitigations include rule/model agreement, confidence thresholds, human review for high-risk classes, certificate audit, and adversarial tests. Critical deployments should treat classifiers as untrusted components and require deterministic policy constraints for high-risk effects.

\subsection{Multi-Intent Tasks}

Users often ask for compound tasks. The intent-certificate layer should avoid issuing one broad lease for the entire conversation. Instead, it can decompose the request into per-step certificates:

\[
C(u)=\{C_1,\ldots,C_n\},
\]

where each \(C_i\) authorizes one effect class and resource bound. If decomposition is uncertain, the gateway should require clarification or draft review.

\paragraph{Operational rule}
Read/summarize certificates may be used to discover candidate objects and present them to the user. Update/delete/export certificates should be issued only after the user confirms the concrete target or approved output scope. One broad certificate should not simultaneously cover search, ranking, and high-risk mutation unless the mutation target is already explicit and bounded.

\section{Related Work}

\subsection{Agent Security Benchmarks}

AgentDojo, InjecAgent, AgentDyn, ToolSandbox, tau-bench, ToolEmu, and MCPAgentBench provide evaluation settings for tool-using agents~\cite{debenedetti2024agentdojo,zhan2024injecagent,li2026agentdyn,lu2024toolsandbox,yao2024taubench,ruan2023toolemu,liu2025mcpagentbench}. These benchmarks primarily measure robustness, task success, or safety under adversarial context. Risk-knowledge experiments further show that a model's ability to recognize risk need not translate into safe action~\cite{tang2025riskknowledge}. The intent layer is complementary: it defines a server-side mechanism that can be evaluated using such tasks.

\subsection{Prompt-Injection Defenses}

StruQ separates prompt and data channels through structured queries and model training~\cite{chen2024struq}. AgentSentry models multi-turn indirect prompt injection as temporal causal takeover and purifies context~\cite{zhang2026agentsentry}. AttriGuard is especially close: it causally attributes a proposed action to trusted user input rather than untrusted observations~\cite{he2026attriguard}. Its action-support signal can feed the downstream authority ceiling, manifest narrowing, payload/effect-bound check, routing outcome, and audit lifecycle of an intent-certificate layer.

\subsection{Tool Selection and MCP Security}

ToolHijacker, ToolTweak, MCPTox, and MCP threat-modeling work show that tool names, descriptions, metadata, and ecosystem composition can create security vulnerabilities~\cite{shi2025toolhijacker,sneh2025tooltweak,wang2025mcptox,huang2026mcptm,shen2026mcp38}. These works motivate intent-aware filtering. Even a perfectly authenticated and well-described tool should not be callable when the user's request does not justify its effect.

\subsection{Agent Authorization and Access Control}

Least privilege, RBAC, ABAC, and OAuth-style delegated authorization remain foundational~\cite{saltzer1975protection,hu1992rbac,nist2014abac,mcp2025authorization}. Usage control adds obligations, conditions, attribute mutation, and ongoing authorization~\cite{park2004ucon}; macaroons make delegated credentials attenuable through contextual caveats~\cite{birgisson2014macaroons}. Security-automata work in turn characterizes the safety policies that execution monitoring can enforce~\cite{schneider2000enforceable}. These mechanisms supply useful policy and enforcement primitives, but do not themselves infer which authority a trusted natural-language request justifies or mediate an untrusted planner at both discovery and action boundaries.

Recent agent-focused work extends OAuth/OIDC delegation with agent credentials and natural-language permission translation, governs inter-agent communication with user-defined policies and cryptographic tokens, enforces intent/context-aware OS policy for computer-use agents, or moves identity- and argument-level policy to an off-host gateway~\cite{south2025authorizedagents,syros2025saga,gong2025csagent,kodathala2026aiauthz}. The mechanism studied here is narrower in principal identity but finer in request lifecycle: it derives a short-lived request envelope, intersects it with an existing gateway policy, filters the model-visible manifest, and rechecks payload/effect bounds before distinguishing draft authority from execution.

Task Shield enforces task alignment against indirect prompt injection; Progent and MiniScope pursue least-privilege tool authorization; CaMeL separates trusted control and untrusted data flows; Prompt Flow Integrity constrains privilege-escalating prompt flows; and AgentSpec provides customizable runtime enforcement~\cite{jia2025taskshield,shi2025progent,zhu2025miniscope,debenedetti2025camel,kim2025promptflow,wang2026agentspec}. These systems are the closest architectural comparators. The mechanism's distinct unit is a server-issued, short-lived request envelope that is intersected with an existing gateway policy both before planning (manifest filtering) and after planning (payload/effect checking), with draft, preflight, confirmation, and execution modeled as different accepted-authority outcomes. Decentralized information-flow control instead attaches principal-specific policies to data propagation and controlled declassification~\cite{myers1997decentralized}; it can isolate untrusted data inside an agent, while the intent layer governs which external tool effects the planner may request. Information-flow isolation, privilege inference, policy languages, and runtime monitors therefore remain complementary controls.

\subsection{Purpose Limitation and Contextual Integrity}

The intent layer is also related to purpose-based access control and contextual privacy. Purpose-based systems restrict data use according to declared purposes rather than only subjects, objects, and operations. Foundational work associates allowed and prohibited purposes with data and enforces them through purpose-aware labeling, conditional roles, and query modification~\cite{byun2008purpose}; recent systems such as Data Guard apply purpose-aware masking in warehouse settings, and purpose limitation has also been studied for data-in-transit systems~\cite{tran2025dataguard,wolf2021purpose}. Contextual integrity frames privacy as appropriate information flow within a social context, using parameters such as sender, recipient, information type, and transmission principle~\cite{nissenbaum2009privacy,barth2006contextual}. Recent assistant work operationalizes contextual integrity for information-sharing and LLM delegation~\cite{ghalebikesabi2024contextualassistants,huang2026needtoknow}.

The mechanism targets request-level purpose restriction for agent side effects. It turns the current user request into a short-lived, auditable purpose-like certificate that narrows an already-authorized agent tool surface; a statically authorized tool remains unavailable when the present request does not justify its effect. Unlike a declared-purpose database policy, this certificate is operational evidence for a bounded action decision, not proof that every downstream use of returned data satisfies a privacy purpose.

\subsection{OpenPort Protocol}

OpenPort Protocol provides the effect-control substrate for this paper~\cite{zhu2026openport}. It defines a governance-first gateway with authorization-dependent discovery, scoped permissions, data policy constraints, draft-first writes, preflight, state witness, idempotency, rate limits, and audit. The intent layer adds an upstream request envelope and explicitly measures the residual authority that effect controls leave in drafts.

\section{Conclusion}

AI-agent access control must evaluate both credential permission and whether the proposed call is justified by the user's current expressed intent. This paper proposed a server-side intent-certificate authorization model with monotone session policy narrowing, intent-aware manifests, and consistency checks between user intent, tool choice, payload, and expected effect. OpenPort provides the existing effect-control substrate: static authorization, policy constraints, draft-first writes, preflight, state witnesses, idempotency, and audit. The intent-certificate layer performs request-level authorization before those effects are reached.

The evidence supports an executable request-envelope mechanism and static-policy non-expansion, while exposing a decisive limitation. In the real-model runtime study, the combined intent-layer and OpenPort path observes 0/198 completed unsafe effects but accepts 40/198 intent-inconsistent draft paths; strict benign completion is 0/108 and 36/108 benign trials produce governed artifacts. The 12-task transfer subset similarly observes 0/24 unsafe completed effects and 5/24 unsafe accepted drafts. Exact certificate bounds and adapter effect contracts remain the principal systems problem.

\appendices

\section{Reference Intent-Layer Reason Codes}

\begin{itemize}
  \item \texttt{agent.intent\_not\_found}
  \item \texttt{agent.intent\_invalid}
  \item \texttt{agent.intent\_binding\_mismatch}
  \item \texttt{agent.intent\_version\_stale}
  \item \texttt{agent.intent\_revoked}
  \item \texttt{agent.intent\_low\_confidence}
  \item \texttt{agent.intent\_conflicting}
  \item \texttt{agent.intent\_tool\_mismatch}
  \item \texttt{agent.intent\_payload\_exceeds\_bound}
  \item \texttt{agent.intent\_expired}
  \item \texttt{agent.intent\_review\_required}
\end{itemize}

\section{Example Target-Profile Certificate}

The following synthetic certificate illustrates the target reference profile used by the formal model. The target includes subject, message, canonicalization, policy, and registry-version bindings beyond the current in-memory prototype.

\begin{lstlisting}[language=json,basicstyle=\ttfamily\scriptsize]
{
  "id": "intent_01",
  "profile": "intent-authorization-reference-v1",
  "requestHash": "sha256:...",
  "subjectBinding": {
    "actorId": "actor_demo",
    "appId": "app_demo",
    "keyId": "key_demo",
    "tenantId": "tenant_demo",
    "workspaceId": "workspace_demo",
    "sessionId": "session_demo",
    "messageId": "message_demo"
  },
  "intentClasses": ["summarize", "read"],
  "deniedIntentClasses": ["create", "update", "delete", "export", "admin"],
  "resourceBounds": {
    "resourceTypes": ["record"],
    "dateRange": { "start": "2026-06-10", "end": "2026-06-17" }
  },
  "effectBounds": {
    "maxRisk": "low",
    "allowExport": false,
    "allowMutation": false
  },
  "confidence": 0.86,
  "maxRoutingOutcome": "execute",
  "lifecycle": {
    "issuedAt": "2026-06-17T23:54:59Z",
    "expiresAt": "2026-06-17T23:59:59Z",
    "status": "active"
  },
  "versions": {
    "canonicalization": "canon-v1",
    "staticPolicy": "policy-demo-17",
    "toolRegistry": "registry-demo-4"
  },
  "classifierProvenance": {
    "source": "hybrid",
    "version": "hybrid-demo-v1"
  },
  "auditDigest": "sha256:..."
}
\end{lstlisting}

\begingroup
\raggedright
\sloppy
\bibliographystyle{IEEEtran}
\bibliography{references}
\endgroup

\end{document}